%% file: main.tex
\documentclass[11pt]{article}
\usepackage{graphicx} 
\usepackage{float}
\usepackage{rotating, graphicx}
\usepackage{booktabs}
\usepackage{multirow}
\usepackage{longtable}
\usepackage{array}
\usepackage{placeins}
\usepackage{caption}
\usepackage{subcaption}
\usepackage[a4paper,left=2cm,right=2cm,top=2.5cm,bottom=2.5cm]{geometry}
\usepackage[natbibapa]{apacite}
\usepackage{xcolor}
\usepackage{url}
\usepackage{etoolbox}
\usepackage{appendix}
\usepackage[english]{babel} 
\usepackage [autostyle, english = american]{csquotes} 

\usepackage{tikz}
\usetikzlibrary{positioning, backgrounds, fit, shapes.arrows, calc}

\MakeOuterQuote{"}
\usepackage{amsmath}
\usepackage{pdflscape}

\usepackage{amssymb}

\usepackage{xcolor}
\definecolor{occblue}{HTML}{273A8F}
\definecolor{occgreen}{HTML}{2C5C34}
\definecolor{occred}{HTML}{B33D3D}
\definecolor{occorange}{HTML}{DE7500}

\newif\ifinappendix
\inappendixfalse

\AtBeginEnvironment{appendices}{\inappendixtrue}
\AtEndEnvironment{appendices}{\inappendixfalse}

\makeatletter
\let\origaddcontentsline\addcontentsline
\renewcommand{\addcontentsline}[3]{%
  \origaddcontentsline{#1}{#2}{#3}%
  \ifinappendix
    \ifstrequal{#1}{toc}{%
      \origaddcontentsline{atoc}{#2}{#3}%
    }{}%
  \fi
}

\newcommand{\tableofappendices}{%
  \begingroup
    \setcounter{tocdepth}{1}
    \@starttoc{atoc}%
  \endgroup
}
\makeatother

\begin{document}

\begin{titlepage}

    \begin{center}
        \Large
        \textbf{Breaking the HISCO Barrier:} \\
        \textbf{Automatic Occupational Standardization with \textit{OccCANINE}*} \\

        \vspace{0.5cm}
        \large
        
        Christian M{\o}ller Dahl, Torben Johansen, Christian Vedel,\\
        \vspace{0.125cm}
         University of Southern Denmark

        \vspace{1cm}
        \textbf{Abstract} \\ 
        \vspace{0.5cm}

        \normalsize
        \parbox{0.90\textwidth}{
            This paper introduces \textit{OccCANINE}, an open-source tool that maps occupational descriptions to HISCO codes. Manual coding is slow and error-prone; \textit{OccCANINE} replaces weeks of work with results in minutes. We fine-tune \textit{CANINE} on 15.8 million description-code pairs from 29 sources in 13 languages. The model achieves 96 percent accuracy, precision, and recall. We also show that the approach generalizes to three systems—OCC1950, OCCICEM, and ISCO-68—and release them open source. By breaking the "HISCO barrier," \textit{OccCANINE} democratizes access to high-quality occupational coding, enabling broader research in economics, economic history, and related disciplines.
            
        }

        \vspace{0.025cm} 
        \parbox{0.90\textwidth}{
            \begin{flushleft}
            \end{flushleft}  
        }
    \end{center}

    \vfill
    
    \footnotesize
    $^*$We want to thank Andreas Slot Ravnholt for conscientious RA work in the early stages of this project. This paper has also benefited from conversations and feedback from Simon Wittrock, Paul Sharp, Matthew Curtis, Casper Worm Hansen, Julius Koschnick, Hillary Vipond, Marco van Leeuwen, and Alexis Litvine. Thanks also to Bram Hilkens who checked 200 Dutch HISCO codes predicted by our model. This project has been presented at the 12th Annual Workshop on ``Growth, History and Development'' (SDU), the PhD seminar series (SDU), the 5th ENCHOS meeting (Linz), the 2nd Norwegian Winter Games in Economic History (Oslo), a Workshop on Machine Learning in Economic History (Lund), the 3rd Danish Historical Political Economy Workshop (SDU), the 2024 EHS Annual Conference (Northumbria University), the 2024 Advanced Methods Workshop (ENS Lyon), the 2024 Annual Meeting of the Danish Econometrics Society (Sandbjerg), the 2025 digHPE Conference (CBS), the 11th International Conference on Computational Social Science (Norrk\"{o}ping), the 2025 Data-Intensive Research Conference (IPUMS), the 8th International Conference on Natural Language and Speech Processing (SDU), the 6th Conference of the European Society of Historical Demography (Bologna), the Applied Economics Seminar Series (University of Copenhagen), and the 2025 SSHA Annual Conference (Chicago). We want to thank every participant at these events for insightful questions and comments. The codebase behind this paper as well as the paper itself benefited from improvements suggested by ChatGPT. \\ 
    All code and guides on how to use \textit{OccCANINE} are available on GitHub \\ \url{https://github.com/christianvedels/OccCANINE}

\end{titlepage}
\newpage


\section{Introduction}

The study of occupational outcomes requires systematic data on people's occupations, and the Historical International Standard Classification of Occupations (HISCO) system has emerged as the standard for categorizing diverse historical occupational data \citep{leeuwen2002hisco}. However, the manual classification of large datasets into HISCO codes has been an arduous and time-consuming process, hampering research progress. A simple back-of-the-envelope exercise illustrates the scale of the problem: Even a highly experienced researcher might spend 10 seconds identifying and typing the correct HISCO code for an occupational description. Thus, coding 10,000 unique descriptions would take around 28 hours of manual labor. In datasets with hundreds of thousands, or even millions, of descriptions, a manual approach quickly becomes infeasible.

In this paper, we present a solution that transforms the task of coding occupations into something that can be completed automatically in minutes or a few hours, including verification of quality. We introduce \textit{OccCANINE}---a small transformer-based language model built on the \textit{CANINE} model by \citet{Clark2022CANINE}---which we finetune on 15.8 million observations of occupational descriptions paired with HISCO codes across 13 languages. This training data was generously contributed by numerous researchers and cover 29 different sources, each of which is cited in Table \ref{tab:training_data}. The resulting model achieves high accuracy, precision, and recall, and can reliably transform raw textual descriptions of occupations into the most appropriate HISCO codes.

The HISCO system was introduced to produce internationally comparable historical occupational data. It, and its various modifications, has since become the most widely used classification scheme for historical occupations, though other schemes remain in use. While this paper focuses on solving the problem of HISCO coding, the method readily extends to other systems. We demonstrate this by also training and releasing \textit{OccCANINE} models for the OCC1950 classification system (used in US historical censuses), the OCCICEM classification system (used in British censuses), and the ISCO-68 classification system (the international system from which HISCO is derived). The lesson is clear: \textit{OccCANINE} provides a general solution to the problem of occupational coding.

By significantly reducing the time and effort required for HISCO coding, our tool democratizes access to historical occupational data. This enables researchers to conduct more extensive and diverse studies while devoting more attention to data quality. This breakthrough has the potential to unlock new insights into occupational trends and shifts over time, contributing valuable knowledge to economics, sociology, political science, history, and related disciplines. Moreover, the paper provides a general recipe for tackling a wide range of similar problems in which large volumes of unstructured text must be mapped into standardized systems. Comparable challenges exist in coding historical customs records, educational descriptions, and many other sources, all of which can be addressed with a similar approach.

The remainder of this paper proceeds as follows. Section~\ref{sec: motivation} motivates our solution. Section~\ref{sec:method} outlines the model architecture, training data, and training procedure. Section~\ref{performance} evaluates performance. Section~\ref{sec:poor_performance} discusses how to use \textit{OccCANINE} in settings where model performance proves inadequate. Section~\ref{sec: conclusion} concludes with recommendations on how to use \textit{OccCANINE}.

\section{Motivation}
\label{sec: motivation}

\subsection{The problem of occupational coding}
\label{sec: motivation-problem}

To generate insights into topics such as women's empowerment \citep{Goldin2006}, social mobility \citep{Clark2023}, the effects of railways \citep{Berger2019Railways, gorges2025tracksmodernityrailroadsgrowth}, first-nature geography \citep{Vedel2023}, the origins of the Industrial Revolution \citep{Allen2009}, and the interplay of technology and development \citep{Morkyr2016}, we need reliable information on people's occupations.
This motivates the collection of large-scale datasets of historical occupational data. Much of this information comes in the form of long lists of textual descriptions (e.g., census records or marriage certificates). An example could be ``\textit{Lives of fishing and farm work},'' which would need to be converted into standardized occupational categories (61110: ``Farmer'' and 64100: ``Fisherman'' in the HISCO system). 

The challenge of transforming raw textual occupational descriptions into standardized categories is not trivial, requiring either extensive manual work by error-prone research assistants or sophisticated methods from the classical natural language processing toolbox. In particular, the diversity of occupational descriptions is problematic.\footnote{For example, the Danish censuses 1787–1901 \citep{LinkLives2022Guide, ddd_method2015} contain no fewer than 17,865 unique descriptions corresponding to the occupation ``farm servant.'' See \cite{Schurer2015} for similar challenges in coding British censuses.} The traditional approach to HISCO coding involves string matching and cleaning using, for instance, regular expressions. Because of negations, changing spelling conventions, typos, and transcription errors, this process quickly becomes complex and error-prone. Typically, the following steps are involved:

\begin{enumerate}
    \item Apply domain knowledge to clean strings: ``Srvnt'' becomes ``servant,'' ``sgt.'' becomes ``sergeant,'' and so on. 
    \item Remove stop words: ``He is the servant'' becomes ``servant''; ``after a long career he retired'' becomes ``retired.''
    \item Manually match the remaining unique strings to the HISCO catalog.
\end{enumerate}

This pipeline must be repeated for every single source, with little scope for generalizability. Our \textit{OccCANINE} model replaces all of these steps and provides significantly greater robustness across data sources.

\subsection{A faster, better, scalable, and replicable solution}

\begin{figure}[ht!]
    \centering
    \caption{Conceptual Model}
    \includegraphics[width=0.5\linewidth]{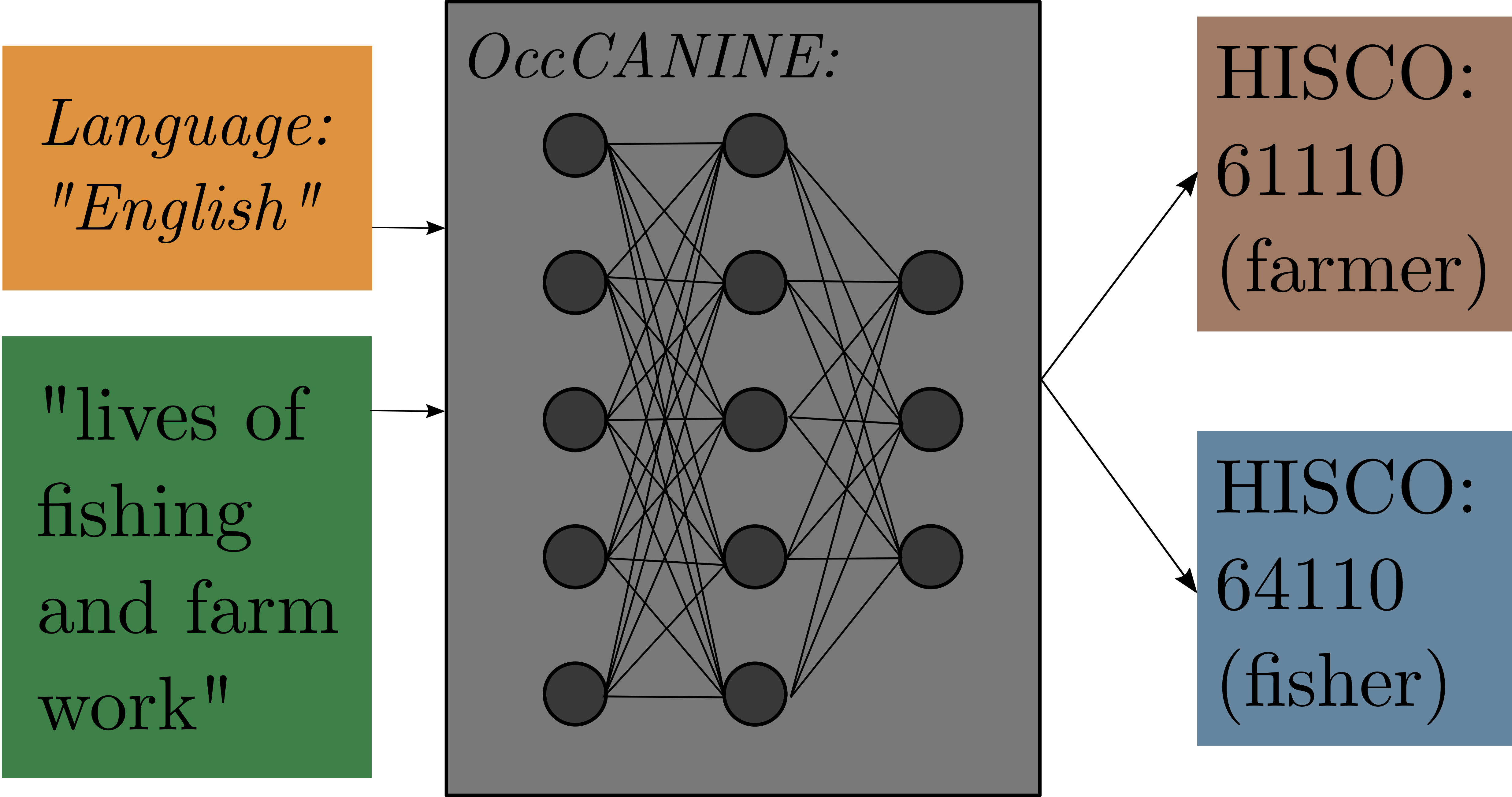}
    \label{fig:concept}
    \parbox{0.5\textwidth}{
        \vspace{0.25cm}
        \footnotesize \textit{Notes}: The figure illustrates the conceptual model: A neural network takes occupational descriptions and (optionally) language as inputs and outputs relevant HISCO codes. 
    }
\end{figure}

The primary barrier is that prior solutions treat occupational coding as surface-level string matching rather than semantic understanding. This root problem makes them slow, inaccurate, unscalable, and unable to generalize across different data sources. Our solution breaks this barrier by treating occupational coding as an end-to-end prediction problem. We train a small language model to understand occupational descriptions as a person would. Specifically, we finetune a preexisting language model on 15.8 million pairs of occupational descriptions and HISCO codes. The resulting model inherently captures the occupational meaning of descriptions, allowing it to process inputs with typos, spelling mistakes, or other irregularities. The model then (similar in spirit to ChatGPT, but smaller) draws on its multilingual knowledge of language and similarity to output the appropriate HISCO code(s). In effect, all the steps described in Section~\ref{sec: motivation-problem} are reduced to a single step: the input is a raw occupational description (and optionally a language), and the output is the relevant HISCO code(s). Figure~\ref{fig:concept} illustrates our approach at a high level, which has the following advantages:

\begin{enumerate}
    \item No need for string cleaning: the text as transcribed can be fed directly into the model.
    \item Accuracy comparable to, or greater than, that of a human labeler.
    \item A general understanding of historical occupations, enabling the model to generalize well to other HISCO coding contexts with little or no fine-tuning. 
    \item Full replicability: given the same inputs, \textit{OccCANINE} will always deliver the same HISCO codes. Replicability has intrinsic scientific value and also reduces variance introduced by human coders. 
\end{enumerate}

\subsection{Literature}
The chronological improvement in performance demonstrates the rapid underlying technological development that now allows the simple method we propose in this paper. To set the stage, we provide a brief overview of recent developments.

A central term in this literature is the \emph{production rate}: the share of occupational descriptions one chooses to automatically transcribe (with the remainder left to a human labeler). This is typically done by only automatically transcribing observations for which a label is assigned with the highest probability (e.g., an 80 percent production rate means that the algorithm produces codes for 80 percent of the samples, while humans produce labels for the remaining 20 percent). Early approaches perform well only at lower production rates. \cite{Patel2012Performance} demonstrate 89 percent agreement with a human labeler at a 71 percent production rate, relying mainly on rule-based, classical NLP approaches. \cite{Gweon2017} propose combining rule-based methods with bag-of-words cosine distance and nearest-neighbors matching. For ``fully automatic labeling'' (100 percent production rate), they achieve 65 percent accuracy on German survey data for the ISCO-88 system (which is similar to the HISCO system), and recommend using the method at lower production rates where performance is higher.\footnote{To achieve 90 percent accuracy, \cite{Gweon2017} require around 40 percent manual labeling.} \cite{Schierholz2020} review this and other machine-learning approaches to automatic occupational labeling. Using boosting trees, they demonstrate around 78 percent agreement at a 100 percent production rate. \cite{turrell2019transforming} propose a heuristics-based method with a production rate of 34 percent for which it agreed 91 percent of the time with a human labeler. \cite{heijden2022} uses features extracted from financial statements and a random forest to predict a standard industry classification, achieving an F1-score of 89 percent.

More recent work builds on Transformers \citep{vaswani2017attention}, including pre-trained models such as BERT \citep{devlin2018bert}. \cite{SuarezGarcia2021} implement a method that combines traditional exact matching and data-cleaning techniques with advanced text analysis (including TF–IDF and Doc2Vec, utilizing BERT) for cases without exact matches. Applied within a conventional machine-learning framework, they report a macro F1-score of 0.65 and a top-5 per-digit macro F1-score of 0.76 in the Canadian National Occupational Classification Scheme. \cite{muhlbach2022occ2vec}, instead of classifying occupations, develops an embedding approach (a high-dimensional semantic representation) from which useful information about any occupation can be extracted. Our method implicitly contains this as well, in the final representation internally in our model before the classification step.

The research most comparable to ours is conducted by \cite{Safikhani2023BERT}, who finetune German BERT and GPT-3 on 47{,}526 observations of German survey data to classify them into the German KldB system. They report a maximum Cohen's kappa of 64.22 percent for the full occupational code in their test data.\footnote{Cohen's kappa is roughly comparable to accuracy but accounts for chance agreement.} 
As a comparison, \cite{Schierholz2020} achieve 48.5 percent Cohen's kappa on the same test data. To the extent that our data are comparable, we outperform these results by a large margin, achieving 96.1 percent accuracy and a 96.3 percent F1-score on our test data. In many cases, this allows us to disregard lower levels of ``production rate,'' as it is barely relevant at this level of performance. Moreover, our method requires no string cleaning, correction of spelling mistakes, or stop-word removal.

\section{Architecture, data, and training procedure}
\label{sec:method}

\subsection{Architecture}

We make use of the \textit{CANINE} architecture \citep{Clark2022CANINE}, a relatively small (127 million parameter) language model based on the transformer architecture \citep{vaswani2017attention}. Modest-sized models often outperform larger models in specialized prediction tasks \citep{bucher2024smallLMstill}, while also offering faster and cheaper inference. \textit{CANINE} was pre-trained on Wikipedia data across 104 languages. Our choice of this architecture is motivated by three factors. \textit{First}, Wikipedia pre-training ensures multilingual capabilities, and it is reasonable to assume some similarity between historical occupational descriptions and Wikipedia text. \textit{Second}, we want the model to be broadly accessible. It is small enough to run (and even finetune modestly) on a common laptop. \textit{Third}, the model uses character-level tokenization. The most commonly used tokenization approaches operate at the wordpiece level, but for archival data this can induce unnecessary model variance. For example, if ``farmer'' is mistyped as ``frmter,'' traditional tokenization may struggle. The \textit{CANINE} architecture is more robust to this type of variation.

Figure~\ref{fig:arch_simple} presents the practical implementation of the conceptual model in Figure~\ref{fig:concept}. We input an occupational description (and optionally a language). \textit{CANINE} (the \textit{``Encoder''}) is then used to obtain a numerical representation of this text string.\footnote{In our default implementation, the encoder produces a high-dimensional representation (a padded sequence of $128 \times 768$ hidden states), which can be summarized into a $768$-dimensional embedding.} These numbers are not directly interpretable, but theoretically, they represent all the information needed to decide on the most relevant HISCO code(s) for the given occupational description. To obtain the HISCO code(s), we have to \textit{decode} this representation. We do this with a \textit{``Decoder''} module that translates the numerical representation into HISCO code(s).

For \textit{OccCANINE}, we implement two decoder modules: (A) a Flat Decoder and (B) a Sequential Decoder.\footnote{In the accompanying \textit{OccCANINE} Python package, these options are exposed via the shorthands \texttt{fast} (Flat) and \texttt{good} (Sequential).} The Flat Decoder is designed for speed and produces a short list of likely HISCO code(s) for each description. The Sequential Decoder, instead, constructs HISCO code(s) digit by digit, which makes it more robust to ambiguity and multi-occupation descriptions out-of-the-box. A practical advantage of the Sequential Decoder is that it predicts one digit at a time, so each step involves a much smaller search space than choosing among all full HISCO codes at once, which makes model training more efficient. 
The Flat Decoder requires that researchers pick a confidence threshold (a cutoff for when a predicted code is accepted), while this is implicitly handled by the Sequential Decoder. For this reason, we generally recommend the Sequential Decoder for most applications, while the Flat Decoder remains an efficient alternative when computational resources are limited.
Appendix~\ref{app:architecture_details} provides the full illustration of the architecture and implementation details, including decoder dimensionalities, the probability-to-prediction mapping for the Flat Decoder, and the decoding procedures (greedy and top-$k$) used with the Sequential Decoder.

\begin{figure}[ht]
    \centering
    \caption{Simplified \textit{OccCANINE} Architecture}
    \input{Figures/arch_simple}
    \label{fig:arch_simple}
    \parbox{0.95\textwidth}{
        \vspace{0.25cm}
        \footnotesize \textit{Notes}: The figure summarizes the practical implementation of \textit{OccCANINE}. We follow a simple Encoder-Decoder architecture. Raw occupational text (optionally with language context) is encoded by \textit{CANINE} and translated into HISCO code(s) by a decoder. We provide two alternative decoders, with separate advantages. 
    }
\end{figure}

\subsection{Data}

This project utilizes training data obtained from public sources or generously provided by fellow researchers, for which we express our gratitude (see Table~\ref{tab:training_data} for a complete list). We process all sources through a common pipeline---character normalization, source-specific quality checks, and validation of HISCO codes---and we additionally augment the data with simple conjunction-based multi-occupation phrases. Appendix~\ref{app:data_processing} summarizes the key steps, and the complete implementation is available in our GitHub repository.

In total, the data consist of 18.5 million observations. Of these, 85 percent (15.8 million) are used for training. The remainder is split across in-training validation (5 percent), post-training validation for model development (5 percent), and final model testing (5 percent). To ensure representativeness across all splits, we stratify by data source. We further conduct out-of-distribution (OOD) evaluation using entirely different data sources.
To improve robustness, we apply additional data augmentation as part of training; Appendix~\ref{app:training_details} provides details.

\begin{table}
    \caption{Training Data}
    \centering
    \small
    \begin{tabular}[t]{l rrr p{6.5cm}}
        \toprule
        Shorthand name & Observations & Language & Share & Source\\
        \midrule
        DK\_census & 5,391,656 & da & 29.08\% & \cite{ddd_method2015}; The Danish National Archives \\
        EN\_marr\_cert & 4,046,203 & en & 21.82\% & \cite{Clark2022ThreeNew} \\
        EN\_uk\_ipums & 3,026,859 & en & 16.32\% & \cite{IPUMS}; Office of National Statistics \\
        SE\_swedpop & 1,793,557 & se & 9.67\% & \cite{SwedPop2022} \\
        JIW\_database & 966,793 & nl & 5.21\% & \cite{DeMoor2021} \\
        EN\_ca\_ipums & 818,657 & unk & 4.41\% & \cite{IPUMS}; Statistics Canada \\
        CA\_bcn & 644,484 & ca & 3.48\% & \cite{BarcelonaHMD2017} \\
        HISCO\_website & 392,248 & mix & 2.12\% &  \cite{HistoryOfWork2024} \\
        SE\_titles & 241,410 & se & 1.30\% & \cite{heikkuri2024technological} \\
        HSN\_database & 184,937 & nl & 1.00\% & \cite{Mandemakers2020} \\
        NO\_ipums & 147,255 & no & 0.79\% & \cite{IPUMS} \\
        FR\_desc & 142,778 & fr & 0.77\% & \cite{HistoryOfWork2024} \\
        EN\_us\_ipums & 139,595 & en & 0.75\% & \cite{IPUMS}; Bureau of the Census \\
        EN\_ship\_data & 103,023 & en & 0.56\% & \citep{schneider2019indefatigable} \\
        EN\_patentee & 94,671 & en & 0.51\% & \cite{Nuvolari2021} \\
        GE\_occupational\_census & 74,168 & ge & 0.40\% & \cite{Albers2023} \\
        EN\_parish & 73,806 & en & 0.40\% & \cite{dePleijt2019} \\
        DK\_cedar & 46,563 & da & 0.25\% & \cite{Ford2023} \\
        SE\_cedar & 45,581 & se & 0.25\% & \cite{SE_CEDAR}\\
        DK\_orsted & 36,608 & da & 0.20\% & \cite{Ford2023} \\
        GE\_Selgert\_Gottlich & 29,159 & ge & 0.16\% & \cite{Gottsel2024} \\
        EN\_oclack & 24,530 & en & 0.13\% & \cite{ipums_cross_walk2017} \\
        EN\_loc & 23,179 & en & 0.13\% & \cite{loc2016} \\
        IS\_ipums & 20,459 & is & 0.11\% & \cite{IPUMS} \\
        SE\_chalmers & 14,426 & se & 0.08\% & \cite{Ford2023} \\
        GE\_ipums & 8,482 & ge & 0.05\% & \cite{IPUMS}; German Federal Statistical Office \\
        GE\_occupations1939 & 5,653 & ge & 0.03\% & Shared by Richard Franke (franker@tcd.ie) \\
        IT\_fm & 4,525 & it & 0.02\% & \cite{Fornasin2016} \\
        EN\_PortArthur & 1,655 & en & 0.01\% & \cite{tuffin2020} \\
        \bottomrule
    \end{tabular}
    \parbox{0.95\textwidth}{
        \vspace{0.25cm}
        \footnotesize
        \textit{Notes}: The table shows a comprehensive overview of the data used for training our model. The shorthand name is the name we use in the remainder of this paper. Observations are the effective number of observations we have after cleaning procedures. The different languages found in the training data are also listed; ``mix'' means that a dataset consisted of multiple languages (explicitly stated for each occupational description) and ``unk'' that the dataset consisted of multiple languages without information available for us to automatically determine the language of each specific occupational description.
        }
    \label{tab:training_data}
\end{table}

\subsection{Training}
The architecture described above has a total of 151 million parameters---when including both encoder and decoder. Model training is performed in random batches of 512 observations over 1{,}605{,}000 steps. That is, the model ``sees'' the full training data around 50 times, and we adjust parameters step by step accordingly. We also apply a few standard regularization and data-augmentation techniques. We use 10 percent dropout \citep{srivastava2014dropout}, random text augmentation, and occasional ``hiding'' of the language label, so the model learns to make predictions even when this information is unavailable. All of this helps produce a more robust model. Appendix~\ref{app:training_details} reports the exact settings.

A limitation of some training data is that occupational codes in the labels are not consistently ordered across samples. For example, ``he fishes and farms'' might be coded as either \textit{[61110: ``General farmer'', 64100: ``Fisherman'']} or \textit{[64100: ``Fisherman'', 61110: ``General farmer'']}. This is not inherently problematic, since our objective is to detect all occupations present in a description regardless of order. However, it poses a challenge for our \textit{Sequential Decoder}, where output order matters. We address this by introducing a novel loss function invariant to the order of predictions. This loss function might be useful for other researchers facing similar problems. See Appendix~\ref{app:order-invariant-loss} for technical details.

\FloatBarrier
\section{Performance} \label{performance}
We evaluate performance on a held-out test set of 931{,}474 observations (5 percent, not used in training) in Table~\ref{tab:performance}. We report accuracy (with partial credit for partial matches), precision, recall, and F1-score. For observations that may have several valid codes, we evaluate precision and recall per observation by comparing the predicted and true sets of codes, compute an F1-score for each observation, and then report the average across all observations; see Appendix~\ref{sec:performance_measures} for details. At the 5-digit level, the Sequential Decoder achieves 96.1 percent accuracy, 96.1 percent precision, 96.7 percent recall, and a 96.3 percent F1-score.\footnote{A natural concern is that high test performance could partly reflect memorization of very common occupational strings that, by sheer chance, appear in both training and test splits (e.g., ``farmer''). Since such strings also occur frequently in genuinely new corpora, removing them understates expected real-world performance. Nevertheless, as a deliberately conservative robustness check, we evaluate on a test subset consisting of strings that never appear in the training data. On this ``guaranteed new strings'' subset, performance remains around 80 percent at the five-digit level across our main metrics, indicating that the model’s accuracy is not driven by overlap in surface forms.}
The Flat Decoder reaches similar performance, but this comes at the cost of requiring threshold tuning. We do this with a grid-search of thresholds from 0.01 to 0.99 on the 926{,}454 validation observations.\footnote{Appendix~\ref{A_optimal_thr} (Table~\ref{tab:A_optimal_thr_flat}) reports language-specific optima. Results here use the globally optimal thresholds that maximize F1-scores.} 

To assess the value of optional language context, we compare \textit{OccCANINE} with and without language information (Table~\ref{tab:performance_by_lang}). Providing language yields a small but consistent improvement; without it, performance remains robust, indicating a multilingual conceptual understanding of occupations. Appendix~\ref{lang_info} (Figure~\ref{fig:with_without_lang}) shows (i) that performance is strong across all 13 trained languages and (ii) the robustness to omitting language information holds across languages.

OccCANINE also achieves relatively fast performance. To benchmark real use cases, we tested our model on 10,000 observations on one of our laptops---mimicking what is available for researchers with a laptop equipped with a GPU. On a Windows machine with an NVIDIA Quadro T2000 4GB GPU, we are capable of predicting 10,000 HISCO codes in 8 minutes and 56 seconds for the Sequential decoder and 1 minute and 38 seconds using the Flat decoder. In comparison, at 10 seconds per observation, this would take a human labeler 27.5 hours.

\begin{table}
    \caption{Test Sample Performance}
    \centering
    \begin{tabular}[t]{llrrrrrr}
        \toprule
        && \multicolumn{5}{c}{Digits} & \\ \cline{3-7}
        Decoder & Statistic & 1 & 2 & 3 & 4 & 5 & Observations \\
        \midrule
        \multirow{4}{*}{\textit{Sequential}}  & Accuracy                   & 0.978 & 0.968 & 0.964 & 0.961 & 0.961 & 931,474\\
                                        & Precision                  & 0.978 & 0.969 & 0.965 & 0.961 & 0.961 & 931,474\\
                                        & Recall                     & 0.982 & 0.975 & 0.971 & 0.968 & 0.967 & 931,474\\
                                        & F1-score                   & 0.979 & 0.971 & 0.967 & 0.963 & 0.963 & 931,474\\
        \midrule
        \multirow{4}{*}{\textit{Flat}}  & Accuracy                   & 0.979 & 0.972 & 0.968 & 0.964 & 0.963 & 931,474\\
                                        & Precision                  & 0.980 & 0.972 & 0.968 & 0.965 & 0.964 & 931,474\\
                                        & Recall                     & 0.985 & 0.980 & 0.978 & 0.975 & 0.975 & 931,474\\
                                        & F1-score                   & 0.981 & 0.975 & 0.971 & 0.968 & 0.968 & 931,474\\
        \bottomrule
    \end{tabular}
    \parbox{0.75\textwidth}{
        \vspace{0.25cm}
        \footnotesize
        \textit{Notes}: The table reports accuracy, precision, recall, and F1-score of our model on the held-out test set. 
        The decoder column indicates which decoding strategy is used. Digit columns refer to the level of granularity: 1 = first digit correct, 2 = first two digits correct, and so on. 
        
    }
\label{tab:performance}
\end{table}

\begin{table}
    \caption{Test Sample Performance With and Without Language Context}
    \centering
    \begin{tabular}{llrrrrr}
        \toprule
        Decoder & Lang context & Accuracy & Precision & Recall & F1-score & Observations \\
        \midrule
        \multirow{3}{*}{\textit{greedy}}  & Yes        & 0.961 & 0.961 & 0.967 & 0.963 & 931,474\\
                                          & No         & 0.958 & 0.958 & 0.965 & 0.960 & 931,474\\ \cmidrule{2-7}
                                          & Difference & 0.003 & 0.003 & 0.003 & 0.003 & 931,474\\
        \midrule
        \multirow{3}{*}{\textit{flat}}    & Yes        & 0.963 & 0.964 & 0.975 & 0.968 & 931,474\\
                                          & No         & 0.960 & 0.961 & 0.974 & 0.965 & 931,474\\ \cmidrule{2-7}
                                          & Difference & 0.004 & 0.004 & 0.001 & 0.003 & 931,474\\
        \bottomrule
    \end{tabular}
    \parbox{0.85\textwidth}{
        \vspace{0.25cm}
        \footnotesize \textit{Notes}: The table shows test-set gains from providing \textit{OccCANINE} with the language of the description (“Lang context = Yes”). Values are computed on \(n=931{,}474\) test observations.
    }
    \label{tab:performance_by_lang}
\end{table}

\FloatBarrier
\textit{OccCANINE} can only learn the patterns provided in the generously contributed training data shown in Table~\ref{tab:training_data}. Yet HISCO coding itself is not free of ambiguity: whether a person belongs to one category or another is sometimes open to interpretation. For example, there are separate codes for ``Policemen and Detectives'' and for ``Policemen and other Maintainers of Law and Order (except Military).'' Our expectation is that, in ambiguous cases, the judgment calls made by human labelers are more often reasonable than not. This would provide more of the \textit{correct} training signal than the \textit{incorrect} one. For this reason, it is particularly interesting to examine whether \textit{OccCANINE} agrees with human labelers from other sources and research projects beyond our training data. We therefore extensively test out-of-distribution (OOD) performance on two distinct groups of corpora. Table~\ref{tab:ood-testing-manual} reports accuracy and agreement rates on samples we manually validated (i.e., with no pre-existing HISCO codes), while Table~\ref{tab:ood-testing-automatic} reports agreement rates with samples that already had HISCO codes assigned (but from sources we did not include in our training data).

The results in Tables~\ref{tab:ood-testing-manual} and \ref{tab:ood-testing-automatic} show that \textit{OccCANINE} achieves high accuracy both in manually checked datasets and in already labeled datasets, despite potential disagreements. For the manually validated cases, we compare the predicted HISCO codes with the definitions and classify them as either (i) strictly correct, (ii) substantially accurate (closely related but not exact), or (iii) incorrect.\footnote{For example, in the Norwegian Biographies we observe an ``engineer and estate owner'' (ingeniør og godseier). \textit{OccCANINE} labels this as ``02000 Engineer, Specialisation Unknown'' and ``61110 General Farmer.'' Since the second code may not be fully appropriate, we classify the result as substantially accurate but not strictly accurate.} Across the two cases shown, \textit{OccCANINE} achieves between 87 and 96 percent accuracy depending on the strictness of the definition.

For the samples that already contained HISCO codes, we use these as the reference. The results are shown in Table~\ref{tab:ood-testing-automatic}, where we report "agreement rates," noting that HISCO is not entirely unambiguous. In some cases, \textit{OccCANINE}’s prediction may be just as valid as the source label, even when the two do not match. Looking at the results, \textit{OccCANINE} performs strongly on the Swedish and Dutch datasets, where ample training data exist. By contrast, performance is lower on the Danish West Indies (mostly English) and UK Bankruptcies sources. A likely explanation is that HISCO codes in these cases were generated with regex-based pipelines, which are brittle and sometimes miss uncommon or compound professions; as a result, genuine occupations were occasionally recorded as $-1$ (``no occupation'') despite this not being appropriate (as we confirm by manual checks). Excluding these cases substantially raises agreement rates (Panel B), suggesting that most residual gaps reflect label coverage rather than model error. For the Italian and German datasets, agreement rates are somewhat lower, which is unsurprising given the limited training data in these languages. Still, the results are encouraging. Even with sparse training material and without any source-specific engineering, \textit{OccCANINE} achieves agreement that appears competitive with the practical alternatives suggested by the literature review, and it provides a useful baseline that can be improved further with light fine-tuning where needed. 

To better understand why \textit{OccCANINE} generalizes across diverse sources, we visualize its encoder embeddings. Encoder embeddings are the model’s internal numerical representations of each occupational description, summarizing the information it uses to predict HISCO codes. When reduced from 768 to two dimensions, occupations cluster naturally by sector, indicating that the model has developed a meaningful semantic representation of historical work (Appendix~\ref{sec:embeddings}).

\begin{table}[h]
    \centering
    \caption{Out-of-Distribution Testing Accuracy: Manual Validation}
    \label{tab:ood-testing-manual}
    \begin{tabular}{l rrr}
        \toprule
        && \multicolumn{2}{c}{Accuracy rate} \\ \cline{3-4}
        Dataset & Observations & Substantial & Strict \\
        \midrule
        Copenhagen Burial Records & 367 & 0.962 & 0.890 \\
        Norwegian Student Biographies & 500 & 0.950 & 0.874 \\
        \bottomrule
    \end{tabular}
    \parbox{0.70\textwidth}{
        \vspace{0.25cm}
        \footnotesize
        \textit{Notes}:  
        The table shows out-of-distribution test performance of our model on two datasets.
        Strict refers to an exact match and substantial to an approximate match.\\  
        \textit{Sources}: Copenhagen Burial Records \citep{LinkLives2022Guide}; Norwegian Student Biographies (sample of fathers occupations in 1831-1920) \citep{FordRanestadSharp2023, HCNC_project}.
    }
\end{table}

\begin{table}[h]
    \centering
    \caption{Out-of-Distribution Testing Accuracy: Automatic Validation}
    \label{tab:ood-testing-automatic}
    \begin{tabular}{l rccccc}
        \toprule
        && \multicolumn{5}{c}{Agreement rate} \\ \cline{3-7}
        Dataset & Observations & 1 & 2 & 3 & 4 & 5 \\
        \midrule
        \multicolumn{5}{l}{\textit{Panel A: Raw data}}  \\
        Swedish Strikes  & 1,430   & 0.957 & 0.946 & 0.927 & 0.897 & 0.896\\
        Dutch Wealth Tax & 200    & 0.938 & 0.892 & 0.780 & 0.767 & 0.760\\
        Italian Marriage Certificates & 26,287 & 0.877 & 0.866 & 0.854 & 0.835 & 0.828\\
        German Denazification Survey  & 800 & 0.785 & 0.731 & 0.665 & 0.647 & 0.638\\
        Danish West Indies & 166,563 & 0.704 & 0.684 & 0.680 & 0.651 & 0.645\\
        UK Bankruptcies & 581,912 & 0.754 & 0.676 & 0.656 & 0.641 & 0.638\\
        \midrule
        \multicolumn{5}{l}{\textit{Panel B: Removed HISCO code '-1': No occupation}}  \\
        Danish West Indies & 101,619 & 0.828 & 0.795 & 0.788 & 0.741 & 0.732\\
        UK Bankruptcies & 326,400 & 0.922 & 0.783 & 0.747 & 0.721 & 0.720\\
        \bottomrule
    \end{tabular}
    \parbox{0.85\textwidth}{
        \vspace{0.25cm}
        \footnotesize
        \textit{Notes}:  
        The table shows out-of-distribution test performance of our model on five datasets.
        1 refers to the first digit being correct, 2 to the first two digits being correct, and so forth.\\
        \textit{Sources}: Swedish Strikes \citep{Enflo2022}; Dutch Wealth Tax (HISCO codes kindly checked by Bram Hilkens) \citep{FamiliegeldRijnland1674}; Italian Marriage Certificates \citep{FreschiMartinez2024}; German Denazification Survey \citep{DeNazDB}; UK Bankruptcies \citep{Korn2024bankruptcy}; Danish West Indies \citep{Galli02024dwi, Ronnback2024dwi}.
    }
\end{table}

We also evaluate model performance across frequency of occupational categories. As expected, accuracy is highest for the most common HISCO codes, while performance gradually declines for rarer ones---a pattern typical in multiclass classification. For the vast majority of codes (the most frequent 99 percent), performance remains very strong, and even for the rarest occupations, average accuracy is still close to 90 percent (Appendix~\ref{A_frequency}). A natural concern is whether this decline in performance might systematically correlate with socio-economic status (SES). Appendix~\ref{A_SES} shows a weak negative association between SES and performance ($\sim$0.05 percentage points lower accuracy per 1-point increase in HISCAM), but this effect is no longer statistically significant once we control for the number of training observations per code. This suggests that the apparent SES effect is driven primarily by the relative scarcity of training data for certain high-SES occupations. In practice, the effect size is negligible, but we nevertheless encourage users to remain attentive to possible systematic biases in sensitive applications. In particular, we recommend plug-and-play approaches based on validation data such as \textit{Prediction Powered Inference} \citep{Angelopoulos2023PPI} or \textit{Design-based Supervised Learning} \citep{egami2024dsl}.

Finally, we demonstrate that the approach is both portable and practical for related problems. Table~\ref{tab:regimes} reports transfer learning performance on three widely used systems---OCCICEM (British censuses), OCC1950 (U.S. historical censuses), and ISCO-68.\footnote{Source: \cite{IPUMS}, Bureau of the Census, Office of National Statistics.} Finetuning on large training sets yields accuracies of 85.8–97.5 percent. A natural concern is that researchers may lack sufficient labeled data or access to large-scale computational resources. To address this, we benchmark finetuning under tighter data and compute budgets. We sample 10,000 observations---an amount that small teams could plausibly label. As shown in Table~\ref{tab:regimes}, Panel B, accuracy remains solid at 80.6–85.4 percent. We also test a variant that freezes the encoder and finetunes only the decoder (Figure~\ref{fig:arch1}), reducing training time to about 20 hours while preserving performance (Table~\ref{tab:regimes}, Panel C). Two conclusions follow: \textit{First}, strong performance is attainable even for classification systems other than HISCO. \textit{Second}, researchers can start with a modest labeled set of 10,000 descriptions, finetune the model, and then use \textit{OccCANINE} to label additional data, which can be corrected and leveraged for iterative improvement.

\begin{table}[H]
    \caption{Model Performance under Different Training Data Constraints}
    \label{tab:regimes}
    \centering
    \small
    \begin{tabular}[t]{llrrrrrl}
\toprule
Target system & Train obs. & Test obs. & Accuracy & Precision & Recall & F1-score & Train time\\
\midrule
\addlinespace[0.3em]
\multicolumn{8}{l}{\textit{Panel A: Full training data}}\\
\hspace{1em}ISCO-68 & 81.4M & 4.8M & 0.938 & 0.938 & 0.992 & 0.958 & $\sim$20 days\\
\hspace{1em}OCCICEM & 68.0M & 4.0M & 0.974 & 0.982 & 0.974 & 0.977 & $\sim$20 days\\
\hspace{1em}OCC1950 & 18.8M & 1.1M & 0.866 & 0.866 & 0.888 & 0.875 & $\sim$20 days\\
\addlinespace[0.3em]
\multicolumn{8}{l}{\textit{Panel B: 10,000 strings}}\\      
\hspace{1em}ISCO-68 & 10,000 & 1,000 & 0.829 & 0.835 & 0.881 & 0.850 & $\sim$48 hours\\
\hspace{1em}OCCICEM & 10,000 & 1,000 & 0.854 & 0.863 & 0.854 & 0.857 & $\sim$48 hours\\
\hspace{1em}OCC1950 & 10,000 & 1,000 & 0.806 & 0.810 & 0.824 & 0.814 & $\sim$48 hours\\
\addlinespace[0.3em]
\multicolumn{8}{l}{\textit{Panel C: 10,000 strings, frozen encoder}}\\
\hspace{1em}ISCO-68 & 10,000 & 1,000 & 0.822 & 0.829 & 0.878 & 0.844 & $\sim$20 hours\\
\hspace{1em}OCCICEM & 10,000 & 1,000 & 0.857 & 0.867 & 0.857 & 0.860 & $\sim$20 hours\\
\hspace{1em}OCC1950 & 10,000 & 1,000 & 0.805 & 0.807 & 0.827 & 0.813 & $\sim$20 hours\\
\bottomrule
\end{tabular}
    \parbox{0.85\linewidth}{
        \vspace{0.25cm}
        \footnotesize
        \textit{Notes}: Train obs. refers to the number of observations used in training. Test obs. to the size of the held-out evaluation sample. Metrics are for the Sequential Decoder only. Training times are measured on a single Tesla V100 (32GB) GPU. Small-sample settings used 102{,}000 training steps.
    }
\end{table}

\begin{figure}[ht!]
    \centering
    \caption{Average Top-\(k\) Improvement}
    \vspace{0.25cm}
    \includegraphics[width=\linewidth]{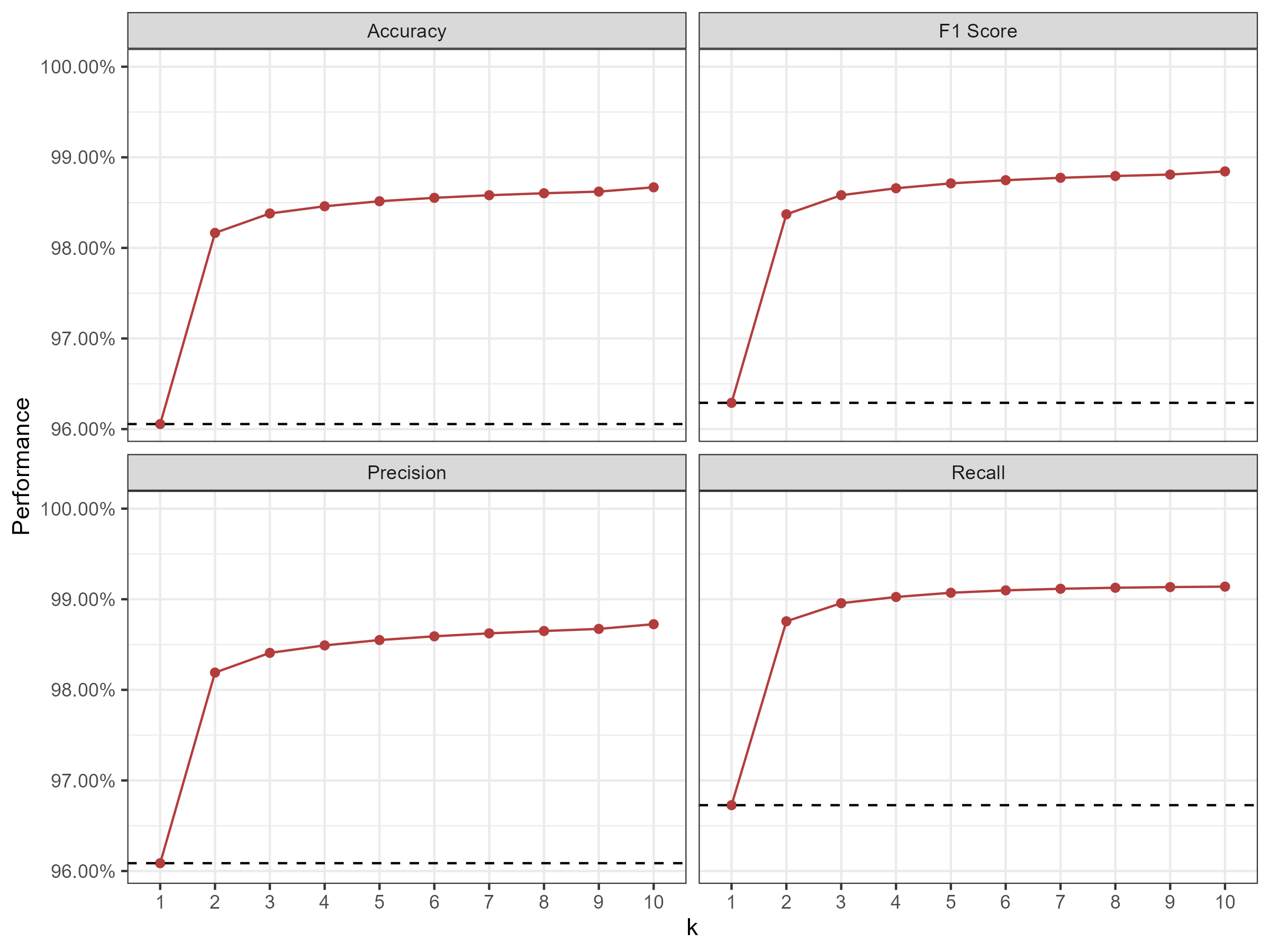}
    \parbox{0.9\textwidth}{
        \small \textit{Notes}: The dashed line marks greedy decoding (\(k=1\), corresponding to the results we report in Table~\ref{tab:performance}); points show performance when evaluation uses the top-\(k\) candidate set. Most gains arrive quickly (especially between \(k=1\) and \(k=2\)), with diminishing returns thereafter. The large jump from \(k=1\) to \(k=2\) indicates that many mistakes are ``near misses.'' Even when the top prediction is wrong, the correct code is often among the highest-ranked alternatives.\\
        \textit{Source}: Test data.
    }
    \label{fig:topk_avg_improvement}
\end{figure}

Beyond fully automatic coding, the Sequential Decoder also supports top-\(k\) decoding, where the model produces a short ranked list of candidate HISCO codes and a human selects the correct one. This is useful when full automation is not appropriate but fast, high-quality standardization is still the goal. Figure~\ref{fig:topk_avg_improvement} summarizes the average pattern: Most gains arrive quickly for small \(k\), with diminishing returns thereafter. 
For example, while the probability of returning the correct HISCO code(s) is around 96 percent, if we allow the model to produce just two suggestions, one of them will be the correct HISCO code(s) in more than 98 percent of cases.
The steep improvement from \(k=1\) to \(k=2\) is also informative about model failures. If errors were typically far off (meaning the correct HISCO code had very low model probability) then adding a second candidate would almost never ``catch'' the truth, and top-\(k\) performance would change little between \(k=1\) and \(k=2\). The fact that performance jumps at \(k=2\) therefore implies that many errors are near misses. The model confuses the correct code with a small set of highly plausible alternatives.

\FloatBarrier

\section{Improving performance}
\label{sec:poor_performance}

Despite generally promising benchmarks, users will encounter applications with materially lower accuracy. This is clear in our out-of-domain evaluations (Tables~\ref{tab:ood-testing-manual} and \ref{tab:ood-testing-automatic}), where domain shift, short or ambiguous strings, and local coding conventions can reduce agreement. Researchers should therefore think explicitly about what level of accuracy is sufficient for their particular use case, and quickly validate performance on at least 100 randomly sampled observations from the target corpus. If that check suggests that performance is inadequate for full automation, \textit{OccCANINE} still offers several practical routes to high-quality assisted coding. This section summarizes a simple escalation strategy.

A first lever is to run at a lower production rate. When error costs are high, it is rarely optimal to auto-code everything. Instead, accept only predictions above a confidence cutoff and route the remainder to manual review. The point is not to hide uncertainty, but to surface it and use it to prioritize human time. Appendix~\ref{A_production_curves} visualizes the trade-off between coverage and performance as the cutoff is tightened. A pragmatic workflow is to run the model once, sort predictions by confidence, validate a small sample around the intended cutoff, and then choose a cutoff that matches either an accuracy target or a review budget.

If the goal is fast, high-quality standardization (rather than full automation), top-\(k\) decoding is a practical middle ground. Rather than committing to a single code, the model proposes the \(k\) most likely candidates (e.g., \(k=2\) or \(k=5\)), which a human can select from. This is particularly useful for low-confidence cases, where the correct code is often among the highest-ranked alternatives even when the top prediction is wrong. Figure~\ref{fig:topk_avg_improvement} illustrates that most gains arrive immediately when moving beyond greedy decoding, with diminishing returns as \(k\) grows. Appendix~\ref{A_production_curves} documents how these gains vary across languages and out-of-domain datasets (Tables~\ref{tab:topk_by_lang} and \ref{tab:topk_ood}, respectively).

When low performance is systematic (e.g., consistent domain shift, a different coding tradition, or a different target classification), thresholding and top-\(k\) can help, but they do not change what the model has learned. In that case, finetuning on a project-specific labeled set is the right solution. Table~\ref{tab:regimes} shows that useful performance is attainable even with a modest labeled sample, and that freezing the encoder can materially reduce training time. A pragmatic approach is to start with a few thousand validated strings, finetune, use the improved model to generate preliminary labels on a larger batch, and then iterate as needed.

\FloatBarrier
\section{Conclusion and recommendations}
\label{sec: conclusion}

We set out to remove a large bottleneck for progress in economic history and related disciplines. Large-scale occupational coding is slow and error-prone to do by hand. By training a small multilingual transformer on 15.8 million labeled description-code pairs, \textit{OccCANINE} delivers around 96\% five-digit accuracy on a held-out test set (and strong out-of-domain performance), turning what would otherwise be weeks of manual work into automated coding in minutes.

For most applications, we recommend using the Sequential Decoder. It achieves performance similar to the Flat Decoder while avoiding explicit threshold selection, which simplifies deployment and reporting. If computational resources are limited, the Flat Decoder remains an efficient alternative. For the Flat Decoder, we recommend a classification threshold of 0.26 to optimize the F1 score and 0.39 to maximize accuracy, based on performance on validation observations. Appendix~\ref{A_optimal_thr} provides language-specific thresholds. A step-by-step guide for applying \textit{OccCANINE} is available in the code repository.

We hope that the freed resources can be used both to scale up research and to improve the quality of work involving standardized historical occupational data. Encouragingly, our method is already powering new and exciting studies.\footnote{Examples include \cite{Vedel2023, andersson2025ascending, chilosi2025smithian, gorges2025tracksmodernityrailroadsgrowth, ford2025origins, Bentzen2024Assimilate, proffit2025linking}.} But we do caution that our method is not perfect. We strongly encourage researchers to validate at least 100 observations and report the achieved accuracy in any publication using our model. This practice enhances transparency, provides readers with a concrete measure of model performance, and underscores the uncertainties inherent in data work. In most cases, we expect that validation checks will confirm the adequacy of the predictions out of the box, or reveal that minor adjustments are sufficient. Where necessary, we also provide a framework for finetuning with a small set of validated observations. We intend to maintain and improve the model, and we encourage anyone using \textit{OccCANINE} to reach out with questions, feedback, and additional data that may help train future versions.

As shown, our approach also transfers well to other classification systems (OCC1950, OCCICEM, ISCO-68), and we suggest that future research extend these efforts. We also see potential for related classification problems, such as coding goods in trade archives, education in biographical sources, or diseases in medical records. While 10,000 observations currently seem a practical starting point for finetuning, future work may also focus on developing more efficient training procedures.

In conclusion, \textit{OccCANINE} represents a significant stride in historical occupational data processing, effectively breaking the HISCO barrier. By automating the translation of occupational descriptions into HISCO codes with high accuracy, our model reduces the costs of working with historical sources and opens new possibilities for research in economics, history, sociology, and beyond.

\vspace{2cm}

\bibliographystyle{apacite}
\bibliography{references}

\newpage
\include{appendix}

\end{document}

%% file: Figures/arch_simple.tex
\begin{tikzpicture}

\tikzset{
  myarrow/.style={
    fill=occblue,
    single arrow,
    minimum width=7mm,
    minimum height=15mm,
    single arrow head extend=1mm
  }
}

\node[fill=occgreen, text=white, inner sep=3pt] (rep) {%
\begin{tabular}{@{}c@{}}
\(\vdots\)\\
\(0.457\)\\
\(0.758\)\\
\(0.024\)\\
\(\vdots\)
\end{tabular}};

\node[
  fit=(rep.north) (rep.south),
  inner sep=2pt,
  right=1mm of rep,
  minimum width=2.4cm,
  label=center:\textbf{Decoder}
] (dec) {};

\node[
  fit=(rep.north) (rep.south),
  inner sep=2pt,
  left=1mm of rep,
  minimum width=2.4cm,
  label=center:{\shortstack{\textbf{Encoder}\\[2pt]\footnotesize\textit{CANINE}}}
] (enc) {};

\begin{scope}[on background layer]
  \fill[occblue!10] (dec.north west)--([yshift=6mm]dec.north east)--
                    ([yshift=-6mm]dec.south east)--(dec.south west)--cycle;
  \fill[occblue!10] (enc.north east)--([yshift=6mm]enc.north west)--
                    ([yshift=-6mm]enc.south west)--(enc.south east)--cycle;
\end{scope}

\node[
  fit=(enc) (dec),
  draw,
  rounded corners,
  inner sep=20pt
] (module) {};

\node[above=1.2mm of module] {\textit{OccCANINE}};

\node[myarrow, right=2mm of dec] (outarr) {};
\node[align=left, right=1.5mm of outarr] (outtxt) {%
\textbf{Prediction}\\
\texttt{61110} General Farmer\\
\texttt{64100} Fisherman};

\node[myarrow, left=2mm of enc] (inarr) {};
\node[align=left, left=1.5mm of inarr] (intxt) {%
\textbf{Text}\\
{\ttfamily ``en[SEP]he}\\
{\ttfamily fishes and}\\
{\ttfamily tends to his}\\
{\ttfamily farm''}
};

\end{tikzpicture}

%% file: appendix.tex
\setcounter{table}{0}
\setcounter{figure}{0}
\renewcommand*{\thesection}{\Alph{section}}
\renewcommand{\thefigure}{A\arabic{figure}}
\renewcommand{\thetable}{A\arabic{table}}
\pagenumbering{roman}

\begin{appendices}

\begin{titlepage}

    \begin{center}
        \Large
        \textbf{Appendix:} \\
        Breaking the HISCO Barrier: \\
        Automatic Occupational Standardization with \textit{OccCANINE} \\

        \vspace{0.5cm}
        \large
        
        Christian M{\o}ller Dahl, Torben Johansen, Christian Vedel,\\
        \vspace{0.125cm}
         University of Southern Denmark

        \vspace{0.5cm}

        \url{https://github.com/christianvedels/OccCANINE}
        
        \vspace{1cm}

        \normalsize
        \begin{minipage}{0.80\textwidth}
    
            \setcounter{tocdepth}{1}
            \tableofappendices

        \end{minipage}
    
        \vfill

    \end{center}

\end{titlepage}

\FloatBarrier
\begin{landscape}
\begin{figure}[h!]
    \centering
    \caption{Architecture of \textit{OccCANINE}}
    \includegraphics[width=\linewidth]{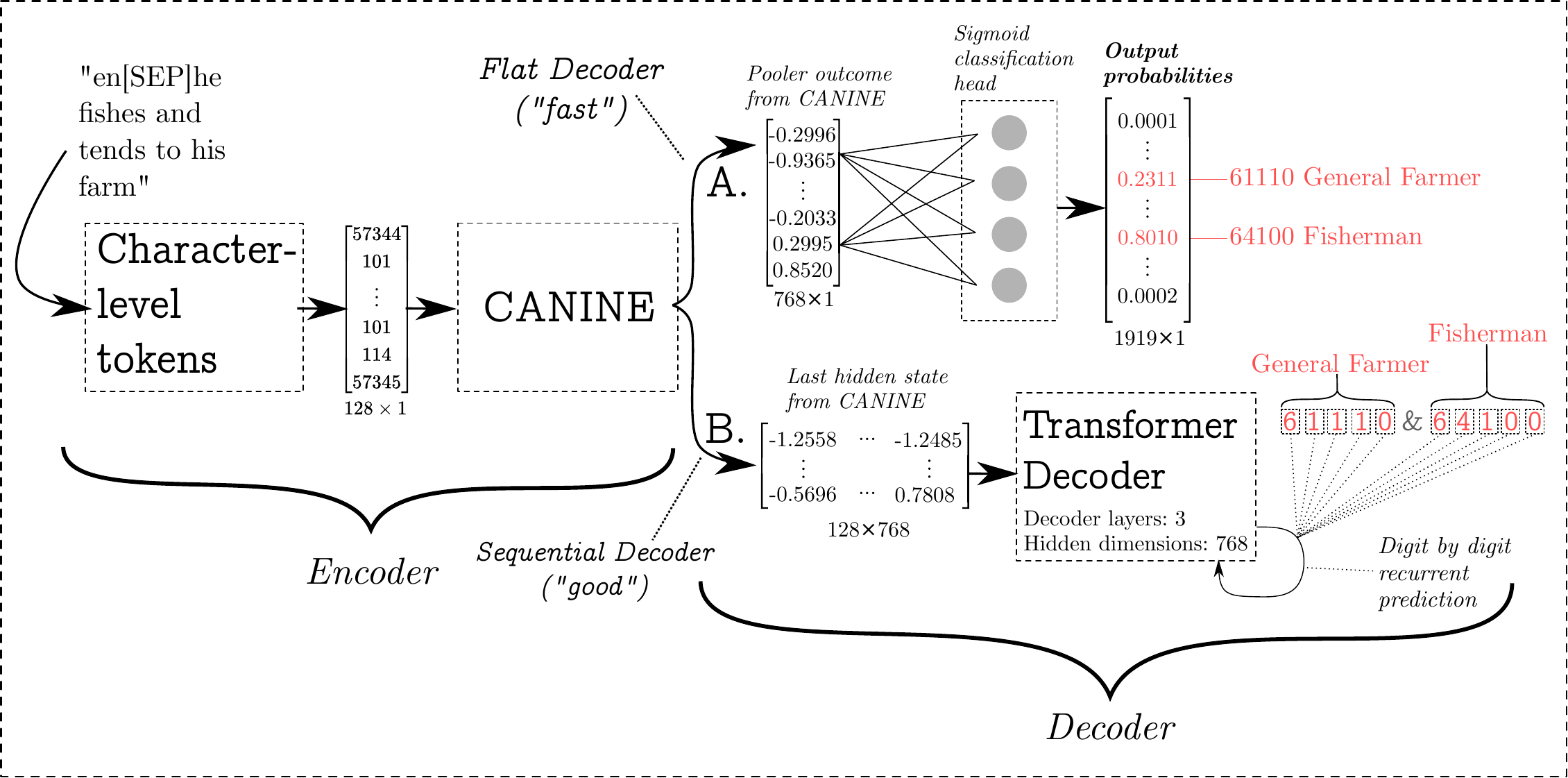}
    \label{fig:arch1}
    \parbox{1.2\textwidth}{
        \vspace{0.25cm}
        \footnotesize \textit{Notes}: The figure shows the architecture of \textit{OccCANINE}, which consists of an encoder and two decoders. In the encoder, input text is converted to character-level tokens, which then serve as input into the \textit{CANINE} model. This is then passed through one of two possible decoders A (denoted \textit{``fast''}) or B (denoted \textit{``good''}). Decoder A is a simple sigmoid classification head, which outputs the probability of each of the 1{,}919 possible HISCO codes. Decoder B is a transformer sequence decoder which predicts the HISCO codes digit by digit. In the example we see that both decoders predict 61110 ``General Farmer'' and 64100 ``Fisherman'', which is a reasonable prediction for the input ``he fishes and tends to his farm''.
    }
\end{figure}
\end{landscape}

\section{Architecture and decoding details}
\label{app:architecture_details}

This appendix provides technical details underlying the architecture described in Section~\ref{sec:method}. The main text gives a broad explanation of what \textit{OccCANINE} does; here we document how predictions are produced with all the technical details.
Figure~\ref{fig:arch1} provides a detailed illustration of the model architecture.

\subsection{Encoder representation}

We use \textit{CANINE} \citep{Clark2022CANINE} as an encoder. Given an input sequence (optionally including language context), the encoder outputs a sequence of hidden states $\mathbf{H} \in \mathbb{R}^{L \times d}$, where $d=768$ in the base \textit{CANINE} model. Occupational descriptions in our corpora are typically short.\footnote{99.97 pct of observations in our training data has string length below 120 characters.} We therefore cap the character-token sequence at $L=128$ (padding shorter inputs and truncating longer ones), yielding a $128 \times 768$ representation per description. When needed we summarize the sequence into $\mathbf{h} \in \mathbb{R}^{768}$ using \textit{CANINE}'s standard pooling method.

\subsection{Flat Decoder (\textit{fast})}

To turn the encoder representation into HISCO codes, one option is a Flat Decoder (``\textit{fast}''). The Flat Decoder is a linear classification head of size $\left[1 \times 1919\right]$---one output for each of the 1,919 potential codes in the HISCO system.\footnote{See \url{https://github.com/cedarfoundation/hisco}.} This output represents the unnormalized score for each HISCO code. We apply sigmoid activation functions to map scores into probabilities and allow multiple HISCO codes to be assigned to a single description. An additional practical advantage of the Flat Decoder is that it returns a full probability vector over all 1{,}919 HISCO codes in a single forward pass, which can be useful for downstream analyses that require uncertainty quantification, ranked candidate lists, or aggregation of predicted probabilities across observations.

To convert probabilities into predicted codes, we apply a threshold. Threshold tuning affects performance and is therefore part of practical deployment. In Section~\ref{performance}, we select recommended thresholds by grid-search on validation data, and Appendix~\ref{A_optimal_thr} reports language-specific optima. In applications where computational resources are limited and speed is paramount, the Flat Decoder is attractive because inference reduces to a single forward pass plus thresholding.

A limitation of the Flat Decoder is that it does not distinguish between cases where multiple codes arise due to ambiguity and cases where they arise because a person had multiple occupations. Moreover, performance depends on the chosen threshold: tightening it improves precision at the cost of recall and coverage, while loosening it does the opposite.

\subsection{Sequential Decoder (\textit{good})}

The second option is a Sequential Decoder (``\textit{good}''). Rather than scoring all HISCO codes independently, the Sequential Decoder predicts HISCO codes digit by digit in a standard sequential transformer decoder architecture. Operationally, the decoder defines a small output vocabulary over digits plus special tokens. In our implementation, the digit vocabulary is $\{-3, -2, -1, 0, 1, \ldots, 9\}$, where special values encode technical placeholders (e.g., start-of-sequence and separators) and the ``no occupation'' label. The decoder predicts the next digit conditional on the encoder representation and previously predicted digits, and we repeat this process to generate up to five HISCO codes per description.

In the main paper, we focus on greedy decoding as described above. But other strategies can be applied to the trained Sequential Decoder. In particular, one can compute the top-$k$ most probable HISCO code sequences for a given input (e.g., $k=5$). This is best understood as assisted coding: the model proposes a short ranked list and a human selects the correct code.

In practice, our library implements top-$k$ decoding as a lightweight extension of greedy decoding. After producing the first (greedy) code, we re-run greedy decoding while masking out the previously returned code, thereby forcing the decoder to take the next-best route through the digit-by-digit search space. Repeating this procedure yields $k$ distinct candidate codes per description without requiring a full beam search or exhaustive enumeration. We rank candidates by their sequence probability, computed as the product of the step-wise conditional probabilities along the generated digit sequence. The resulting list is typically most useful precisely in the cases where greedy decoding is uncertain: many errors are ``near misses,'' and the correct code is often among the next few highest-ranked alternatives.

\FloatBarrier
\section{Optimal threshold} \label{A_optimal_thr}

\begin{figure}
    \centering
    \caption{Optimal Threshold}
    \vspace{0.5cm}
    \includegraphics[width=\linewidth]{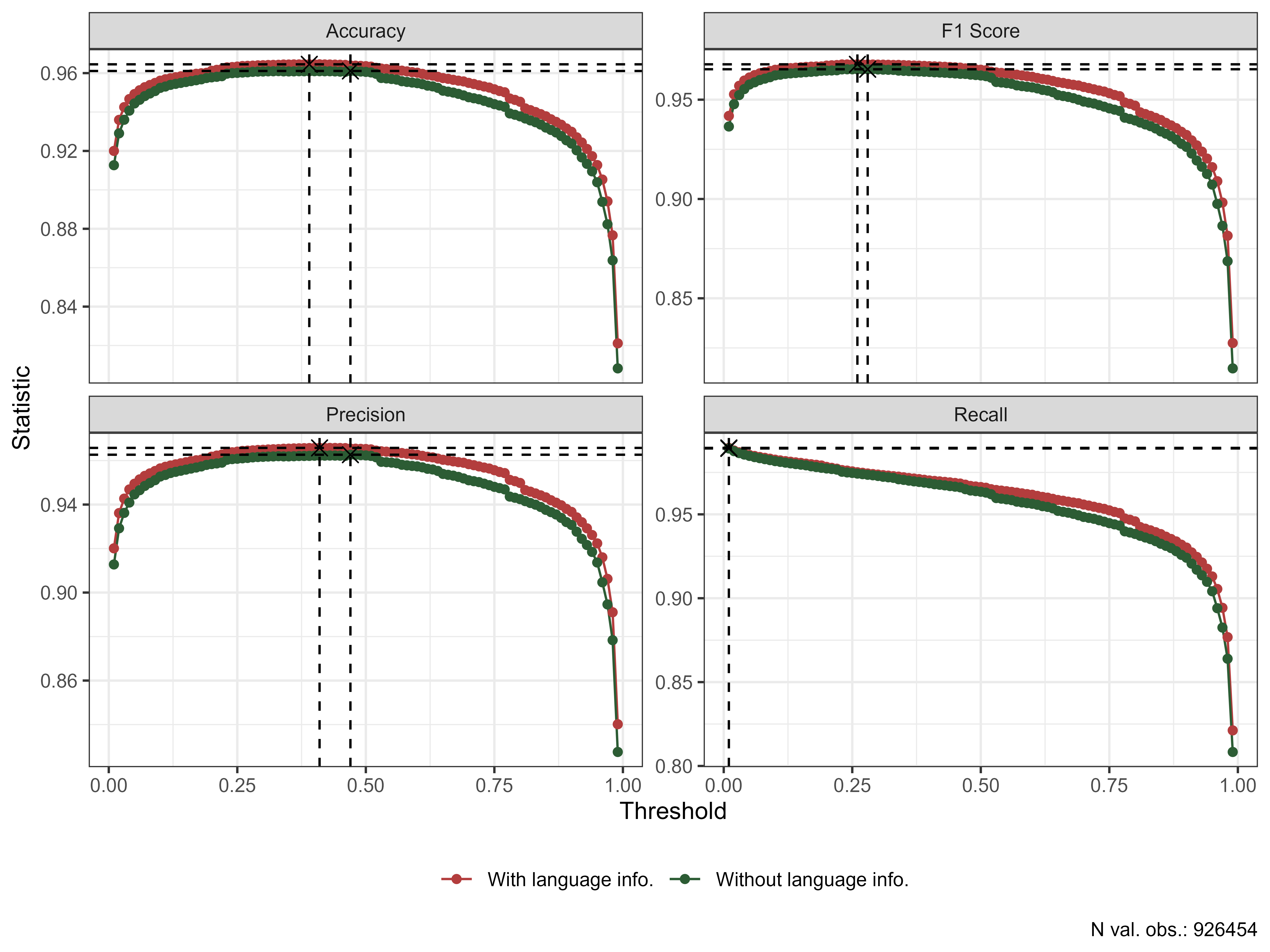}
    \parbox{0.9\textwidth}{
            \small \textit{Notes:} Model performance at various classification thresholds, depicting accuracy, F1 score, precision, and recall. The red line represents performance with language information and the green line without language information. The dashed, vertical lines indicate the optimal thresholds for each metric. The figure shows results for the \textit{flat} decoder.
        }
    \label{fig:Thr_and_perf}
\end{figure}

Table~\ref{tab:best_thr} reports pooled validation performance for the \textit{flat} decoder with and without language identifiers and, importantly, the corresponding pooled test performance obtained by applying the validation-tuned threshold to held-out test predictions. In practice, pooled test performance is virtually identical to pooled validation performance, with occasional differences only in the last reported digit, consistent with minor sampling variation rather than systematic shifts. Supplying language information yields small but consistent gains across metrics. Figure~\ref{fig:Thr_and_perf} visualizes the validation metrics across threshold values for the \textit{flat} decoder; vertical dashed lines mark the validation-optimal thresholds reported in Table~\ref{tab:best_thr}.

Table~\ref{tab:A_optimal_thr_flat} lists metric-specific optimal thresholds by language for the \textit{flat} decoder, selected on each language's validation split and evaluated on the corresponding test split at the validation-tuned threshold. These language-level thresholds provide recommended starting points when applying \textit{OccCANINE} in language-specific settings using the \textit{flat} decoder.

\begin{table}
\caption{Best Overall Performance by Language Context (Flat Decoder Only)}
\centering
\begin{tabular}[t]{llrrr}
\toprule
Metric & Lang. info. & Optimal thr. & Validation & Test\\
\midrule
 & No  & 0.47 & 0.961 & 0.961\\
\cmidrule{2-5}\nopagebreak
\multirow{-2}{*}{\raggedright\arraybackslash Accuracy} & Yes & 0.39 & 0.965 & 0.964\\
\cmidrule{1-5}
 & No  & 0.28 & 0.965 & 0.965\\
\cmidrule{2-5}\nopagebreak
\multirow{-2}{*}{\raggedright\arraybackslash F1 score} & Yes & 0.26 & 0.968 & 0.968\\
\cmidrule{1-5}
 & No  & 0.47 & 0.963 & 0.962\\
\cmidrule{2-5}\nopagebreak
\multirow{-2}{*}{\raggedright\arraybackslash Precision} & Yes & 0.41 & 0.966 & 0.966\\
\cmidrule{1-5}
 & No  & 0.01 & 0.989 & 0.989\\
\cmidrule{2-5}\nopagebreak
\multirow{-2}{*}{\raggedright\arraybackslash Recall} & Yes & 0.01 & 0.990 & 0.990\\
\bottomrule
\end{tabular}
\vspace{0.5cm}
\\
\parbox{0.7\textwidth}{
  \small \textit{Notes:} Peak out-of-sample performance of \textit{OccCANINE} across pooled validation predictions, summarized by whether explicit language information was provided to the model (\textit{Lang.~info.}). Values are maxima of each metric over a 0.01--0.99 threshold grid (step 0.01); "Optimal thr." is the threshold attaining that maximum on the validation set. "Test" reports performance on the pooled test set evaluated at the validation-tuned threshold. Validation results are based on \(n=926{,}454\) observations and test results on \(n=931{,}474\). See Table~\ref{tab:A_optimal_thr_flat} for language-specific thresholds.
}
\label{tab:best_thr}
\end{table}

\FloatBarrier

\FloatBarrier
\newpage
\begingroup
\small 
\centering
\begin{longtable}[t]{lrrlrrr}
\caption{Optimal Thresholds by Language (Flat Decoder)}\\
\label{tab:A_optimal_thr_flat} \\
\toprule
Language & N val obs. & N test obs. & Statistic & Optimal thr. & Validation & Test\\
\midrule
\endfirsthead
\caption[]{Optimal thresholds by language (flat decoder) (continued): }\\
\toprule
Language & N val obs. & N test obs. & Statistic & Optimal thr. & Validation & Test\\
\midrule
\endhead

\endfoot
\bottomrule
\endlastfoot

 &  &  & Accuracy & 0.31 & 0.997 & 0.997\\
\cmidrule{4-7}\nopagebreak
 &  &  & F1 score & 0.30 & 0.998 & 0.998\\
\cmidrule{4-7}\nopagebreak
 &  &  & Precision & 0.69 & 0.998 & 0.999\\
\cmidrule{4-7}\nopagebreak
\multirow{-4}{*}{\raggedright\arraybackslash ca} & \multirow{-4}{*}{\raggedleft\arraybackslash 32,565} & \multirow{-4}{*}{\raggedleft\arraybackslash 33,029} & Recall & 0.01 & 0.999 & 0.999\\
\cmidrule{1-7}\pagebreak[0]

 &  &  & Accuracy & 0.41 & 0.990 & 0.990\\
\cmidrule{4-7}\nopagebreak
 &  &  & F1 score & 0.29 & 0.991 & 0.991\\
\cmidrule{4-7}\nopagebreak
 &  &  & Precision & 0.52 & 0.991 & 0.991\\
\cmidrule{4-7}\nopagebreak
\multirow{-4}{*}{\raggedright\arraybackslash da} & \multirow{-4}{*}{\raggedleft\arraybackslash 274,180} & \multirow{-4}{*}{\raggedleft\arraybackslash 274,557} & Recall & 0.01 & 0.997 & 0.997\\
\cmidrule{1-7}\pagebreak[0]

 &  &  & Accuracy & 0.41 & 0.940 & 0.940\\
\cmidrule{4-7}\nopagebreak
 &  &  & F1 score & 0.26 & 0.945 & 0.945\\
\cmidrule{4-7}\nopagebreak
 &  &  & Precision & 0.41 & 0.941 & 0.941\\
\cmidrule{4-7}\nopagebreak
\multirow{-4}{*}{\raggedright\arraybackslash en} & \multirow{-4}{*}{\raggedleft\arraybackslash 377,142} & \multirow{-4}{*}{\raggedleft\arraybackslash 378,783} & Recall & 0.01 & 0.983 & 0.983\\
\cmidrule{1-7}\pagebreak[0]

 &  &  & Accuracy & 0.30 & 0.963 & 0.954\\
\cmidrule{4-7}\nopagebreak
 &  &  & F1 score & 0.30 & 0.971 & 0.964\\
\cmidrule{4-7}\nopagebreak
 &  &  & Precision & 0.75 & 0.977 & 0.974\\
\cmidrule{4-7}\nopagebreak
\multirow{-4}{*}{\raggedright\arraybackslash es} & \multirow{-4}{*}{\raggedleft\arraybackslash 433} & \multirow{-4}{*}{\raggedleft\arraybackslash 427} & Recall & 0.01 & 0.995 & 0.999\\
\cmidrule{1-7}\pagebreak[0]

 &  &  & Accuracy & 0.32 & 0.918 & 0.917\\
\cmidrule{4-7}\nopagebreak
 &  &  & F1 score & 0.21 & 0.932 & 0.932\\
\cmidrule{4-7}\nopagebreak
 &  &  & Precision & 0.80 & 0.948 & 0.948\\
\cmidrule{4-7}\nopagebreak
\multirow{-4}{*}{\raggedright\arraybackslash fr} & \multirow{-4}{*}{\raggedleft\arraybackslash 14,401} & \multirow{-4}{*}{\raggedleft\arraybackslash 14,528} & Recall & 0.01 & 0.990 & 0.990\\
\cmidrule{1-7}\pagebreak[0]

 &  &  & Accuracy & 0.33 & 0.886 & 0.881\\
\cmidrule{4-7}\nopagebreak
 &  &  & F1 score & 0.21 & 0.903 & 0.901\\
\cmidrule{4-7}\nopagebreak
 &  &  & Precision & 0.75 & 0.918 & 0.912\\
\cmidrule{4-7}\nopagebreak
\multirow{-4}{*}{\raggedright\arraybackslash ge} & \multirow{-4}{*}{\raggedleft\arraybackslash 6,681} & \multirow{-4}{*}{\raggedleft\arraybackslash 6,720} & Recall & 0.01 & 0.974 & 0.971\\
\cmidrule{1-7}\pagebreak[0]

 &  &  & Accuracy & 0.19 & 0.952 & 0.920\\
\cmidrule{4-7}\nopagebreak
 &  &  & F1 score & 0.19 & 0.965 & 0.940\\
\cmidrule{4-7}\nopagebreak
 &  &  & Precision & 0.69 & 0.967 & 0.930\\
\cmidrule{4-7}\nopagebreak
\multirow{-4}{*}{\raggedright\arraybackslash gr} & \multirow{-4}{*}{\raggedleft\arraybackslash 86} & \multirow{-4}{*}{\raggedleft\arraybackslash 88} & Recall & 0.01 & 1.000 & 0.989\\
\cmidrule{1-7}\pagebreak[0]

 &  &  & Accuracy & 0.20 & 0.971 & 0.964\\
\cmidrule{4-7}\nopagebreak
 &  &  & F1 score & 0.17 & 0.974 & 0.968\\
\cmidrule{4-7}\nopagebreak
 &  &  & Precision & 0.20 & 0.971 & 0.964\\
\cmidrule{4-7}\nopagebreak
\multirow{-4}{*}{\raggedright\arraybackslash is} & \multirow{-4}{*}{\raggedleft\arraybackslash 1,054} & \multirow{-4}{*}{\raggedleft\arraybackslash 1,042} & Recall & 0.01 & 0.989 & 0.981\\
\cmidrule{1-7}\pagebreak[0]

 &  &  & Accuracy & 0.32 & 0.998 & 0.997\\
\cmidrule{4-7}\nopagebreak
 &  &  & F1 score & 0.32 & 0.999 & 0.997\\
\cmidrule{4-7}\nopagebreak
 &  &  & Precision & 0.32 & 1.000 & 0.997\\
\cmidrule{4-7}\nopagebreak
\multirow{-4}{*}{\raggedright\arraybackslash it} & \multirow{-4}{*}{\raggedleft\arraybackslash 223} & \multirow{-4}{*}{\raggedleft\arraybackslash 244} & Recall & 0.01 & 1.000 & 1.000\\
\cmidrule{1-7}\pagebreak[0]

 &  &  & Accuracy & 0.24 & 0.978 & 0.976\\
\cmidrule{4-7}\nopagebreak
 &  &  & F1 score & 0.22 & 0.980 & 0.979\\
\cmidrule{4-7}\nopagebreak
 &  &  & Precision & 0.34 & 0.979 & 0.978\\
\cmidrule{4-7}\nopagebreak
\multirow{-4}{*}{\raggedright\arraybackslash nl} & \multirow{-4}{*}{\raggedleft\arraybackslash 58,795} & \multirow{-4}{*}{\raggedleft\arraybackslash 59,536} & Recall & 0.01 & 0.993 & 0.992\\
\cmidrule{1-7}\pagebreak[0]

 &  &  & Accuracy & 0.34 & 0.987 & 0.985\\
\cmidrule{4-7}\nopagebreak
 &  &  & F1 score & 0.34 & 0.988 & 0.986\\
\cmidrule{4-7}\nopagebreak
 &  &  & Precision & 0.63 & 0.988 & 0.986\\
\cmidrule{4-7}\nopagebreak
\multirow{-4}{*}{\raggedright\arraybackslash no} & \multirow{-4}{*}{\raggedleft\arraybackslash 7,982} & \multirow{-4}{*}{\raggedleft\arraybackslash 8,188} & Recall & 0.01 & 0.996 & 0.996\\
\cmidrule{1-7}\pagebreak[0]

 &  &  & Accuracy & 0.27 & 0.993 & 0.989\\
\cmidrule{4-7}\nopagebreak
 &  &  & F1 score & 0.27 & 0.995 & 0.993\\
\cmidrule{4-7}\nopagebreak
 &  &  & Precision & 0.45 & 0.998 & 0.996\\
\cmidrule{4-7}\nopagebreak
\multirow{-4}{*}{\raggedright\arraybackslash pt} & \multirow{-4}{*}{\raggedleft\arraybackslash 1,011} & \multirow{-4}{*}{\raggedleft\arraybackslash 976} & Recall & 0.01 & 1.000 & 1.000\\
\cmidrule{1-7}\pagebreak[0]

 &  &  & Accuracy & 0.35 & 0.969 & 0.969\\
\cmidrule{4-7}\nopagebreak
 &  &  & F1 score & 0.21 & 0.972 & 0.972\\
\cmidrule{4-7}\nopagebreak
 &  &  & Precision & 0.52 & 0.972 & 0.972\\
\cmidrule{4-7}\nopagebreak
\multirow{-4}{*}{\raggedright\arraybackslash se} & \multirow{-4}{*}{\raggedleft\arraybackslash 110,995} & \multirow{-4}{*}{\raggedleft\arraybackslash 111,877} & Recall & 0.01 & 0.986 & 0.986\\
\cmidrule{1-7}\pagebreak[0]

 &  &  & Accuracy & 0.39 & 0.985 & 0.984\\
\cmidrule{4-7}\nopagebreak
 &  &  & F1 score & 0.38 & 0.986 & 0.986\\
\cmidrule{4-7}\nopagebreak
 &  &  & Precision & 0.39 & 0.985 & 0.984\\
\cmidrule{4-7}\nopagebreak
\multirow{-4}{*}{\raggedright\arraybackslash unk} & \multirow{-4}{*}{\raggedleft\arraybackslash 40,906} & \multirow{-4}{*}{\raggedleft\arraybackslash 41,479} & Recall & 0.01 & 0.998 & 0.998\\

\end{longtable}

\begin{center}
  \parbox{0.82\textwidth}{%
    \small \textit{Notes:} Language-specific optimal classification thresholds for the \textit{flat} decoder. Thresholds are tuned on the validation split by maximizing each metric over a 0.01--0.99 grid (step 0.01). \textit{Validation} reports the metric value at its validation-optimal threshold, and \textit{Test} reports performance on the corresponding test split evaluated at that same validation-tuned threshold. In practice, test performance closely mirrors validation performance, with any differences typically confined to the last reported digit.
  }
\end{center}
\endgroup

\FloatBarrier
\section{Performance with and without language} \label{lang_info}

\begin{figure}[h]
    \centering
    \caption{Performance With and Without Language Context}
    \vspace{0.5cm}
    \includegraphics[width=\linewidth]{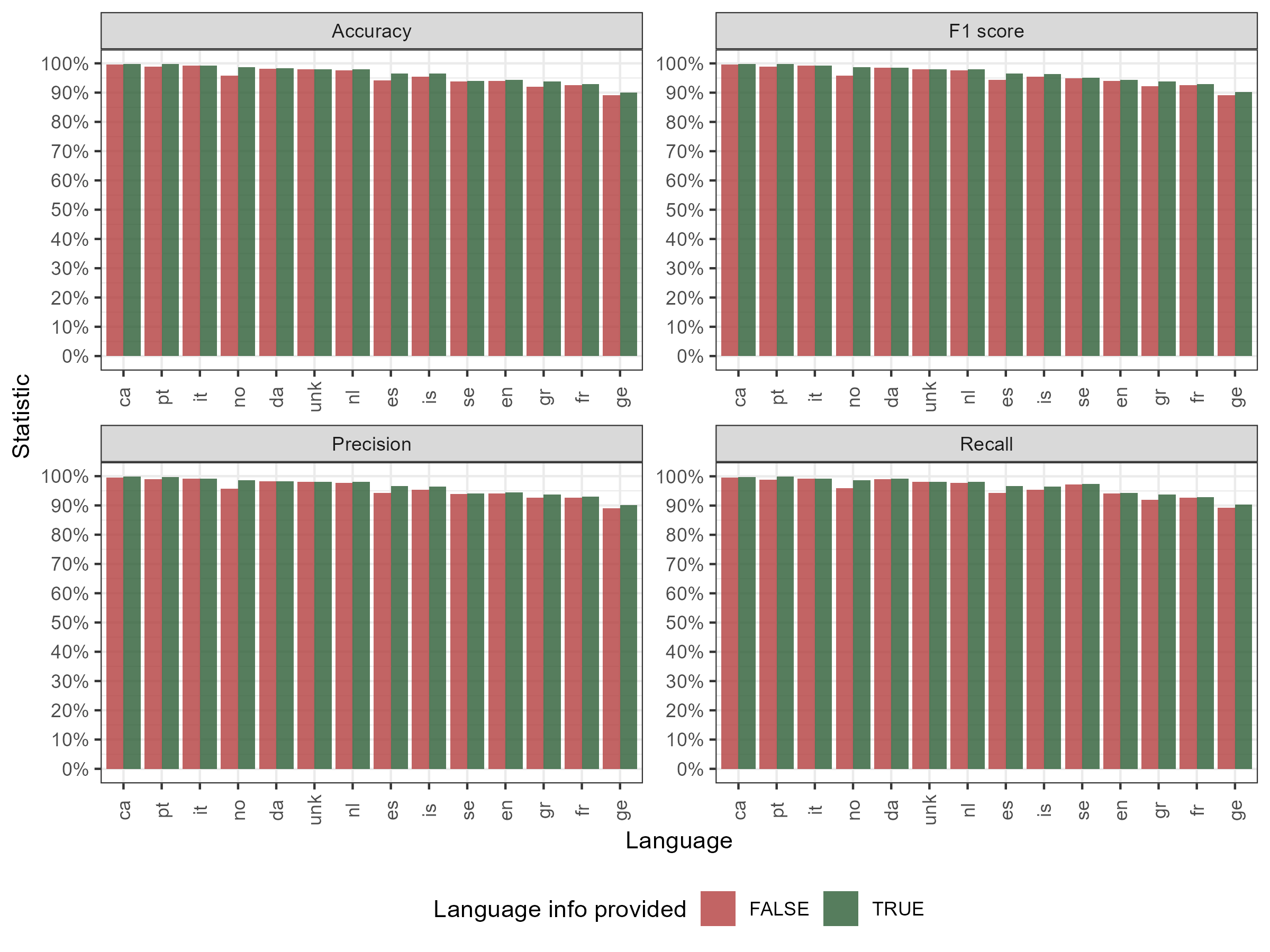}
    \parbox{0.9\textwidth}{
            \small \textit{Notes:} Performance metrics when including/excluding language information. Accuracy, F1 score, precision, and recall included. Each bar represents a language.  
        }
    \label{fig:with_without_lang}
\end{figure}

Figure~\ref{fig:with_without_lang} indicates the benefits of providing language information: it yields modest improvements for all languages.

\FloatBarrier
\section{Label frequency and performance} \label{A_frequency}
\begin{figure}[htbp]
    \centering
    \caption{Performance for each HISCO Code by Frequency}    
    \begin{subfigure}[t]{0.75\textwidth}
        \centering
        \includegraphics[width=\linewidth]{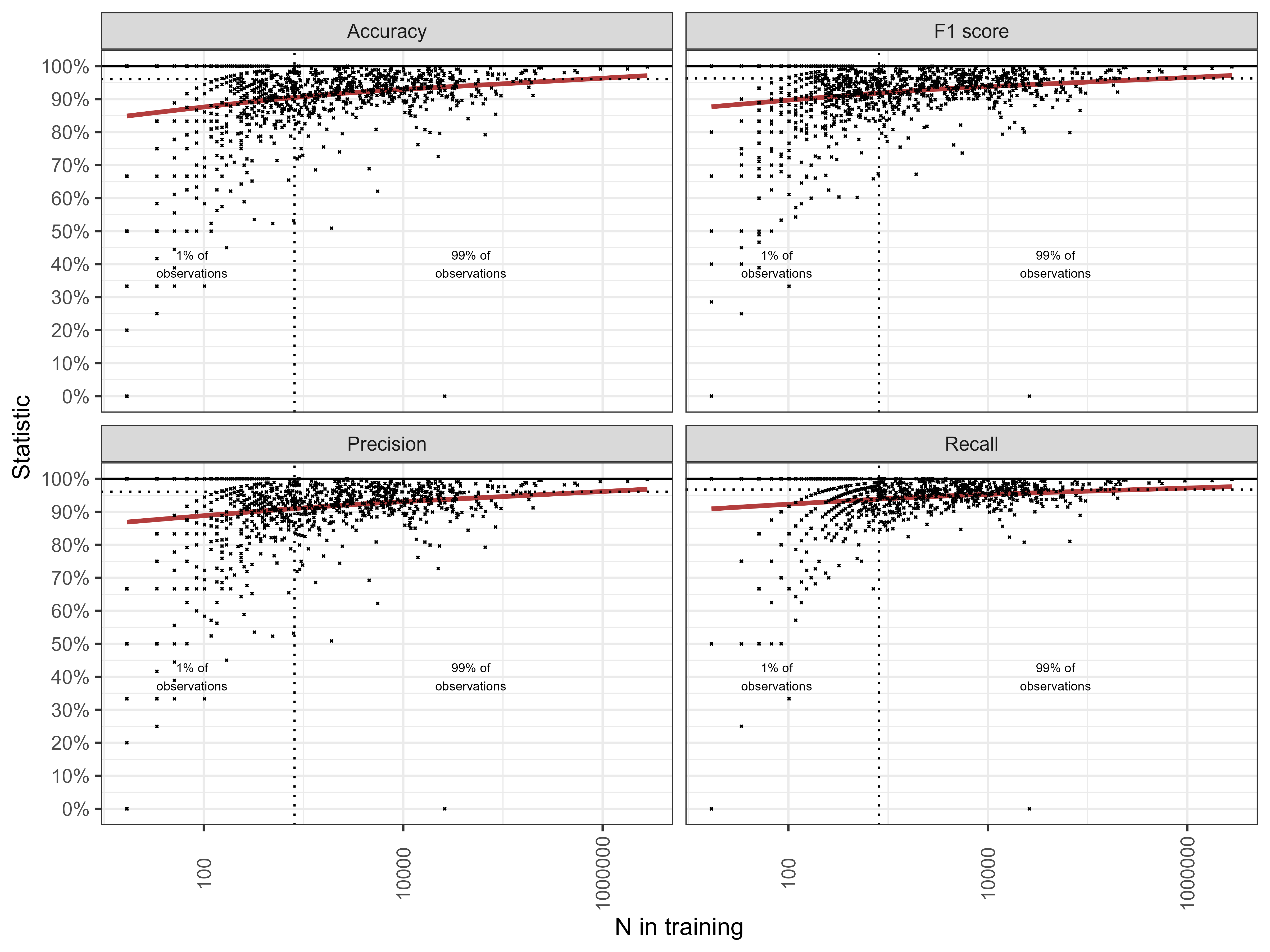}
        \caption{\textit{Flat decoder} results}
        \label{fig:flat_performance_by_frequency}
    \end{subfigure}
        \hfill
    \begin{subfigure}[t]{0.75\textwidth}
        \centering
        \includegraphics[width=\linewidth]{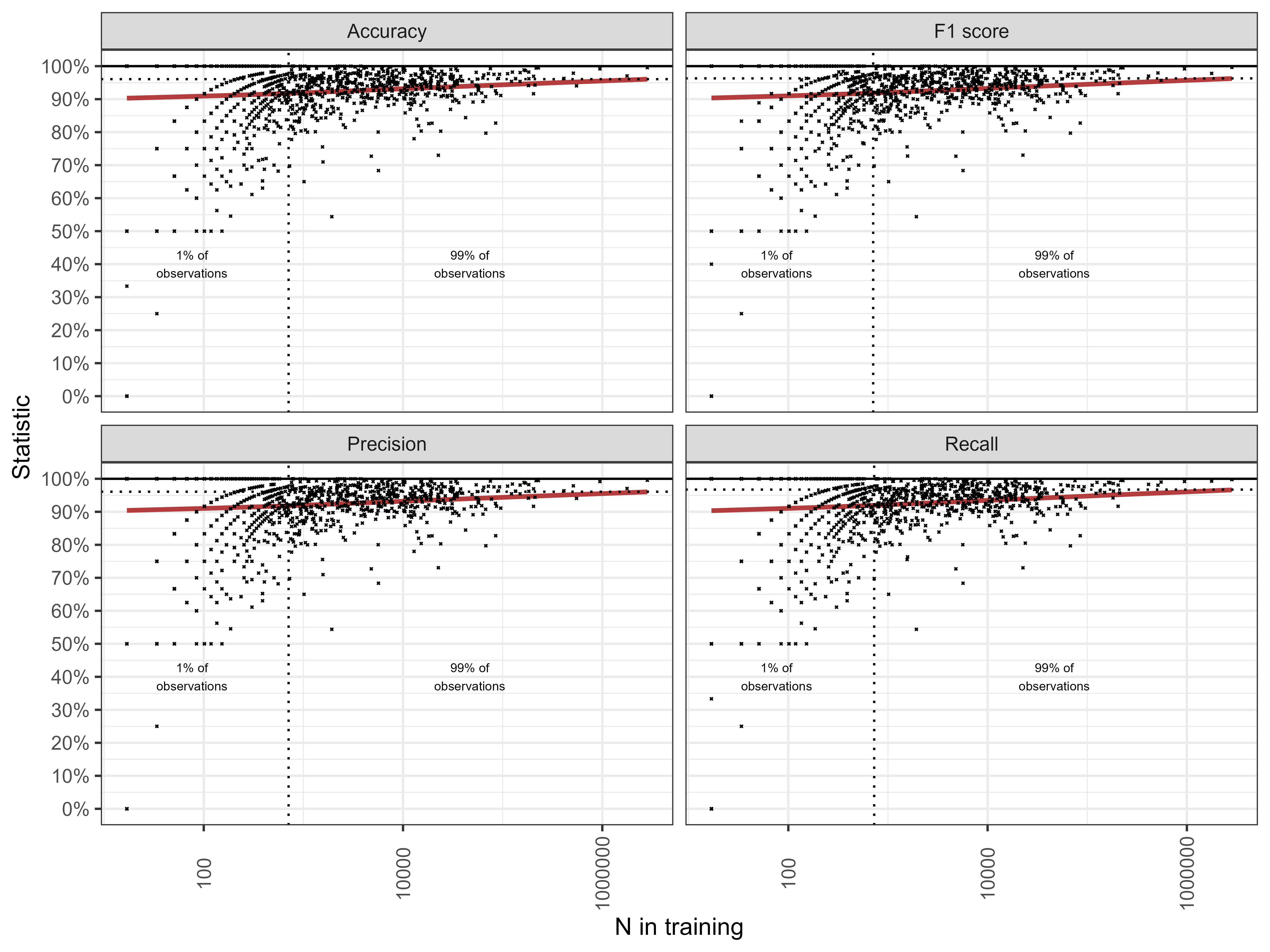}
        \caption{\textit{Sequential decoder} results}
        \label{fig:greedy_performance_by_frequency}
     \end{subfigure}
    \vspace{0.5cm}
    \parbox{0.9\textwidth}{
                \small \textit{Notes:}
                The model's performance stratified by HISCO codes in terms of accuracy, precision, recall, and F1 score. Each point represents a HISCO code, with the position along the x-axis indicating its frequency in the training data. The red line indicates a smoothed trend across the data points. \\
                \textit{Source:} Test data for metrics, training data for frequencies.
            }
    \label{fig:Performance_by_hisco}
\end{figure}

To study whether the performance of our model differs across sectors and/or jobs, Figure \ref{fig:Performance_by_hisco} shows the performance of our model by HISCO code.
It reveals that while the model performs well across the board, there is a tendency for the error rate to increase for rarer occupations, as is common in any multiclass classification problem. For illustrative purposes, we divide these classes into the 99th percentile and the 1st percentile according to their relative share in the training data. 
We also include a trend line estimated by a generalized additive model in Figure \ref{fig:Performance_by_hisco}, which indicates that HISCO codes that are underrepresented in the training data (shown by the vertical lines indicating the 1 percent and 99 percent cumulative frequency of observations) tend to have lower performance metrics.
To account for this variation in performance, users may consider further finetuning with deliberate oversampling of rare occupations for better performance in particular occupations of interest. For the vast majority of HISCO codes observed, we see strong performance, while even below this threshold, performance is still on average around 90 percent.

\FloatBarrier
\section{Performance by SES} \label{A_SES}

\begin{figure}[h]
    \centering
    \caption{Model Performance and SES}
    \begin{subfigure}[t]{0.75\textwidth}
        \centering
        \includegraphics[width=\linewidth]{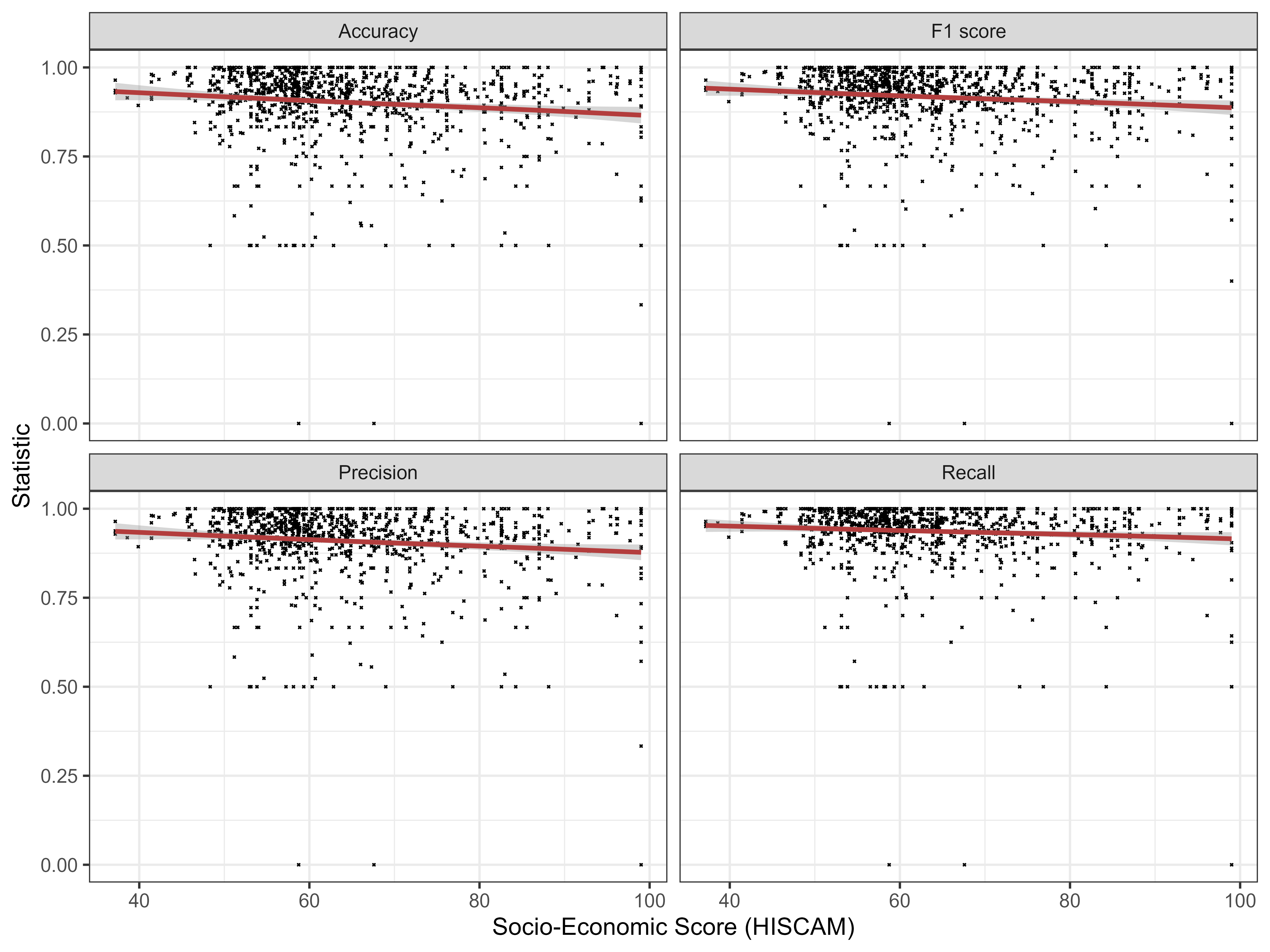}
        \caption{\textit{Flat decoder} results}
        \label{fig:flat_performance_by_ses}
    \end{subfigure}
        \hfill
    \begin{subfigure}[t]{0.75\textwidth}
        \centering
        \includegraphics[width=\linewidth]{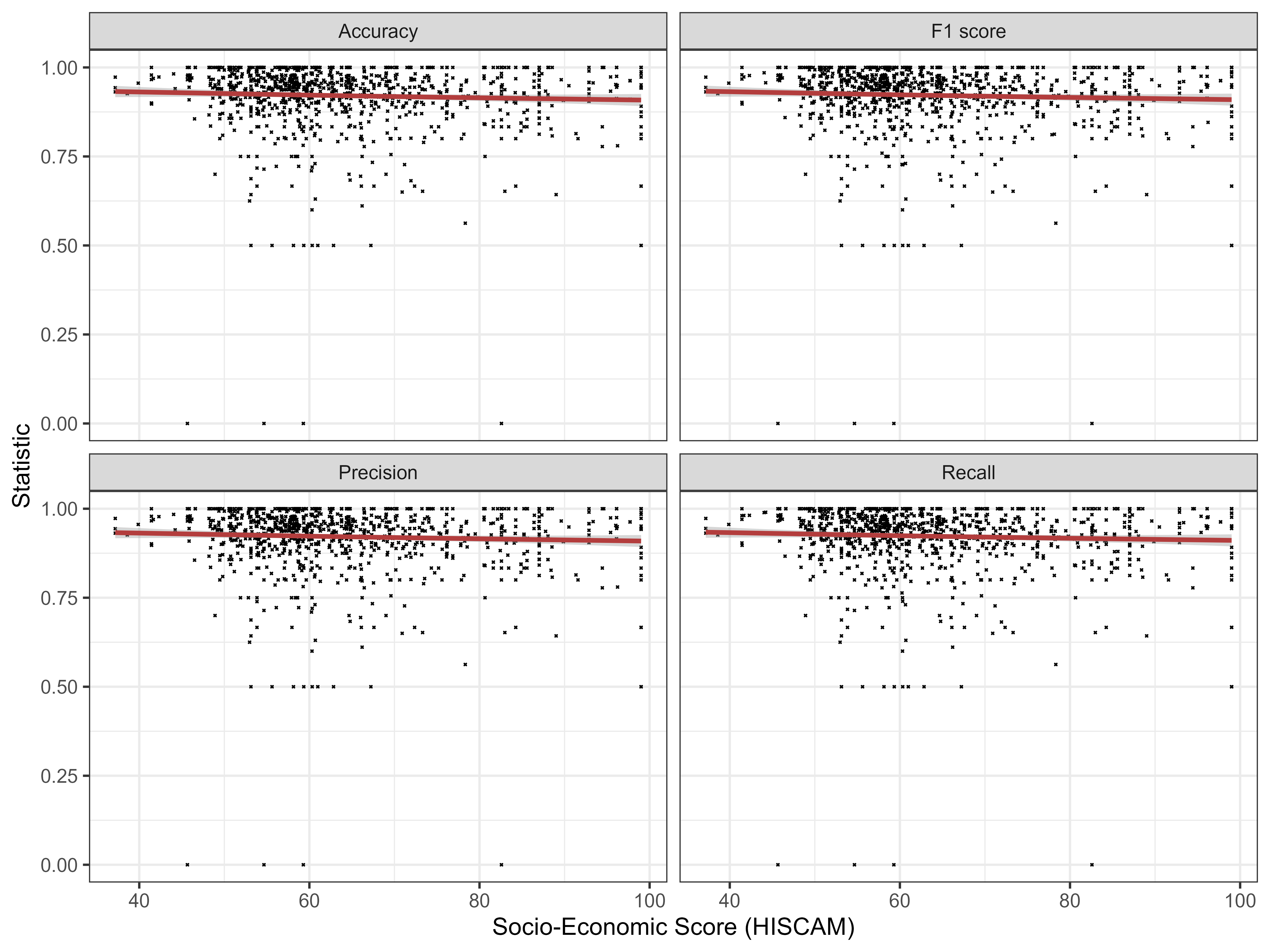}
        \caption{\textit{Sequential decoder} results}
        \label{fig:greedy_performance_by_ses}
     \end{subfigure}
    \parbox{0.9\textwidth}{
        \small \textit{Notes:} The scatter plots depict the relationship between model performance (accuracy, precision, recall, and F1 score) and HISCAM socio-economic scores. Each point corresponds to a HISCO code, with the red line indicating a smooth trend across the data points. No discernible pattern suggests a non-biased performance of the model across different socio-economic strata. \\
        \textit{Source:} Test data; HISCAM scores from \citet{hiscoR}.
        }
    \label{fig:Performance_by_ses}
\end{figure}

\begin{table}[h]
\centering
\caption{Performance Metrics and Socio-economic Status}
\small
\begin{tabular}{lcccc}
   \tabularnewline \midrule \midrule
                         & Accuracy & F1 score & Precision & Recall\\  
                         & (1)      & (2)      & (3)       & (4)\\  
   \toprule
   \multicolumn{5}{c}{\textit{Panel A: SES and performance}} \\
   SES value             & -0.0005$^{*}$ & -0.0005$^{*}$ & -0.0005$^{*}$ & -0.0005$^{*}$\\   
                         & (0.0003)      & (0.0003)      & (0.0003)      & (0.0003)\\   
   \toprule
   \multicolumn{5}{c}{\textit{Panel B: Controlling for training obs.}} \\
   SES value             & -0.0004       & -0.0004       & -0.0004       & -0.0004\\   
                         & (0.0003)      & (0.0003)      & (0.0003)      & (0.0003)\\   
   $\log(n)$             & 0.0045$^{**}$ & 0.0046$^{**}$ & 0.0044$^{**}$ & 0.0049$^{***}$\\   
                         & (0.0018)      & (0.0018)      & (0.0018)      & (0.0018)\\   
   \midrule
   Observations          & 1,055         & 1,055         & 1,055         & 1,055\\  
   \midrule \midrule
\end{tabular}
\parbox{0.57\textwidth}{
                
    \small \textit{Notes:} This table presents the correlation between the socioeconomic score of an HISCO code (HISCAM) and the model performance for that HISCO code. The coefficients of the HISCAM score are small across all specifications. This is also the case when controlling for the logarithm of the number of observations in Panel B. The table shows results for the \textit{greedy} decoder, but qualitatively equivalent results can be shown for the \textit{flat} decoder as well. Heteroskedasticity-robust standard errors in parenthesis. *** $p< 0.01$ ** $p< 0.05$ * $p< 0.10$.  \\
    \textit{Source:} Test data; HISCAM scores from \citet{hiscoR}.
            }
\label{tab:reg_ses}
\end{table}

In historical occupational data, the rarity of certain occupations could correlate with the socio-economic status (SES) associated with the occupation, and in turn, the performance of our model might systematically vary with the socioeconomic status of occupations.\footnote{Note that this problem might also affect squishy wet neural networks; also known as human labellers.} This potentially introduces systematic bias when using our method in applied settings. To investigate this, we first visualize the relationship between the SES, derived from HISCO codes using the HISCAM score,\footnote{HISCAM assigns each HISCO code a value between 1 and 99 \citep{Lambert2013}.} and the model's performance metrics. This plot, shown in Figure \ref{fig:Performance_by_ses}, allows us to examine if there is any correlation between SES values and accuracy, precision, recall, or F1 score. 

The trend line in Figure \ref{fig:Performance_by_ses} is estimated using a generalized additive model, which imposes no linearity, but we end with a remarkably linear relationship for which reason we also find it reasonable to run simple linear regressions to test the relationship. The results from these regressions reveal a small negative effect ($p<0.10$), suggesting that \textit{OccCANINE}'s performance is weakly correlated with the socio-economic status implied by the occupational codes. The interpretation of the regression coefficients is that a one point increase in socio-economic score would decrease performance by 0.05 percentage-points.

Table \ref{tab:reg_ses} shows the results of these regression, with Panel A showing the simple linear regression of our performance metrics on HISCAM and Panel B controlling for sample size.

We suspect that this behavior stems from \textit{OccCANINE} seeing fewer rare occupations in training, and these have higher socio-economic scores on average. To test this mechanisms we include the log of the number of training observations for each HISCO code (Panel B). After the inclusion of this, the otherwise borderline significant coefficient becomes insignificant. Both panels show qualitatively the same result: There is practically no correlation between HISCAM and the performance of \textit{OccCANINE}. Nevertheless, we urge users to be vary of any systematic errors when using our method. 

\section{Production curves and top-\(k\) decoding}
\label{A_production_curves}

Top-\(k\) decoding turns automatic coding into assisted coding. Instead of committing to a single occupational code, the model returns a short ranked list of the \(k\) most likely codes. A human coder then chooses the correct code from this shortlist (or rejects the list and codes manually). In practice, this is often the most efficient way to use a model when the goal is high-quality standardization: the model does the search, while the human makes the final judgment call.

The results in this section document three broad patterns. \textit{First}, top-\(k\) decoding shifts performance-coverage trade-offs in a favorable direction: for a given production rate, it typically increases the share of correct suggestions, and the gains are largest in the low-confidence tail. \textit{Second}, the marginal value of adding more candidates declines quickly. Most of the benefit comes from moving beyond greedy decoding to $k = 2$, while very large \(k\) adds comparatively little. \textit{Third}, the gains are heterogeneous: they vary across languages and are especially pronounced in out-of-domain settings, where errors under greedy decoding are often near misses rather than complete failures.

\begin{figure}
    \centering
    \caption{Production Curves by Confidence Threshold}
    \vspace{0.5cm}
    \includegraphics[width=\linewidth]{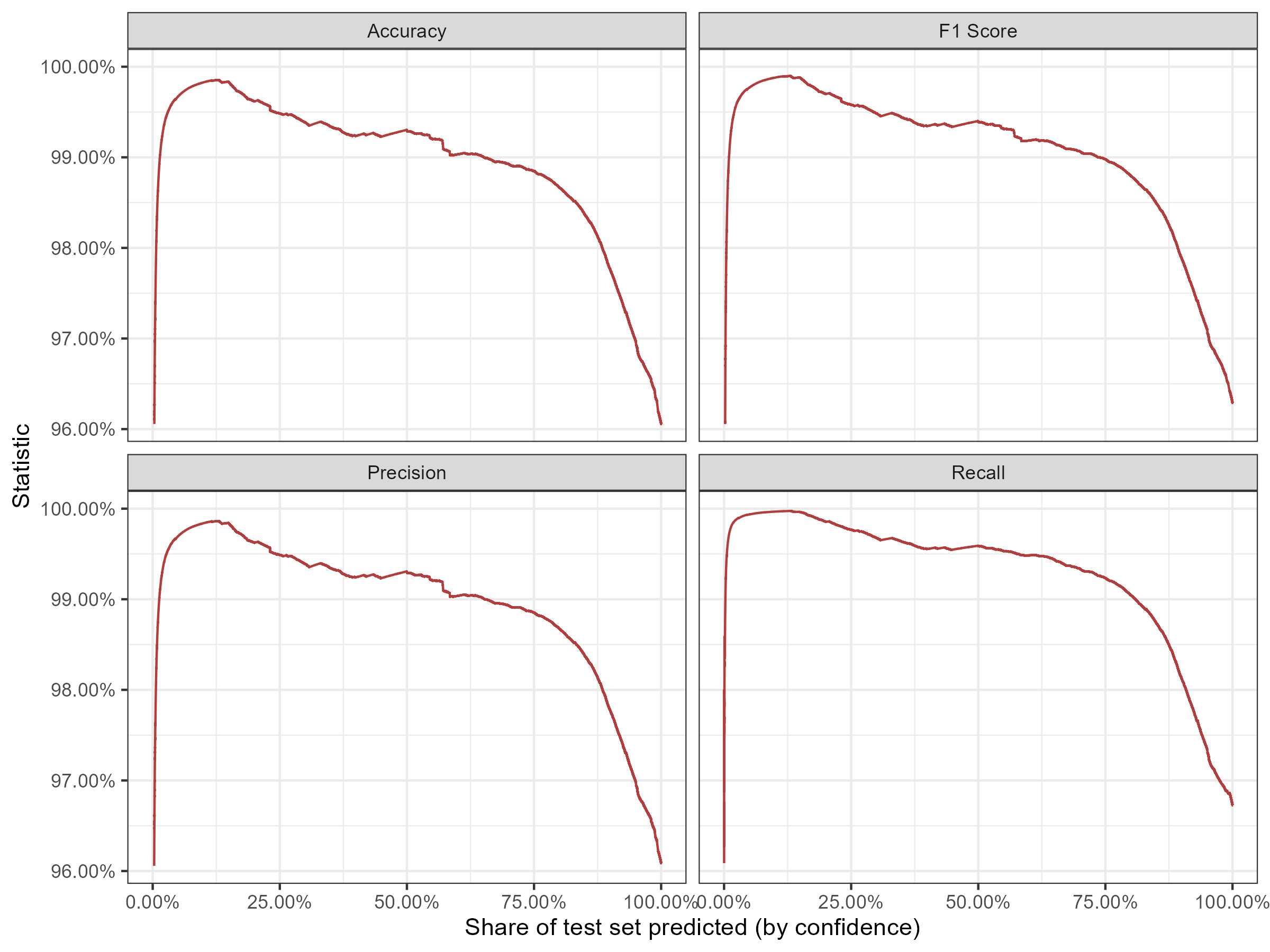}
    \parbox{0.9\textwidth}{
        \small \textit{Notes:} 
        The x-axis shows the share of observations for which a prediction is produced after applying a confidence cutoff and the y-axis shows performance among predicted observations.
        Tightening the confidence cutoff increases performance among auto-coded observations at the cost of lower coverage/production rate. \\
        \textit{Source}: Test data.
    }
    \label{fig:production_curve}
\end{figure}

Figure~\ref{fig:production_curve} shows how performance improves as one tightens a confidence cutoff and accepts fewer automatic predictions. This curve is useful as a diagnostic. it makes explicit how much performance can be bought by lowering coverage, and it identifies where the model is confident but wrong versus simply uncertain.

\begin{figure}
    \centering
    \caption{Production Curves with Top-\(k\) Decoding}
    \vspace{0.5cm}
    \includegraphics[width=\linewidth]{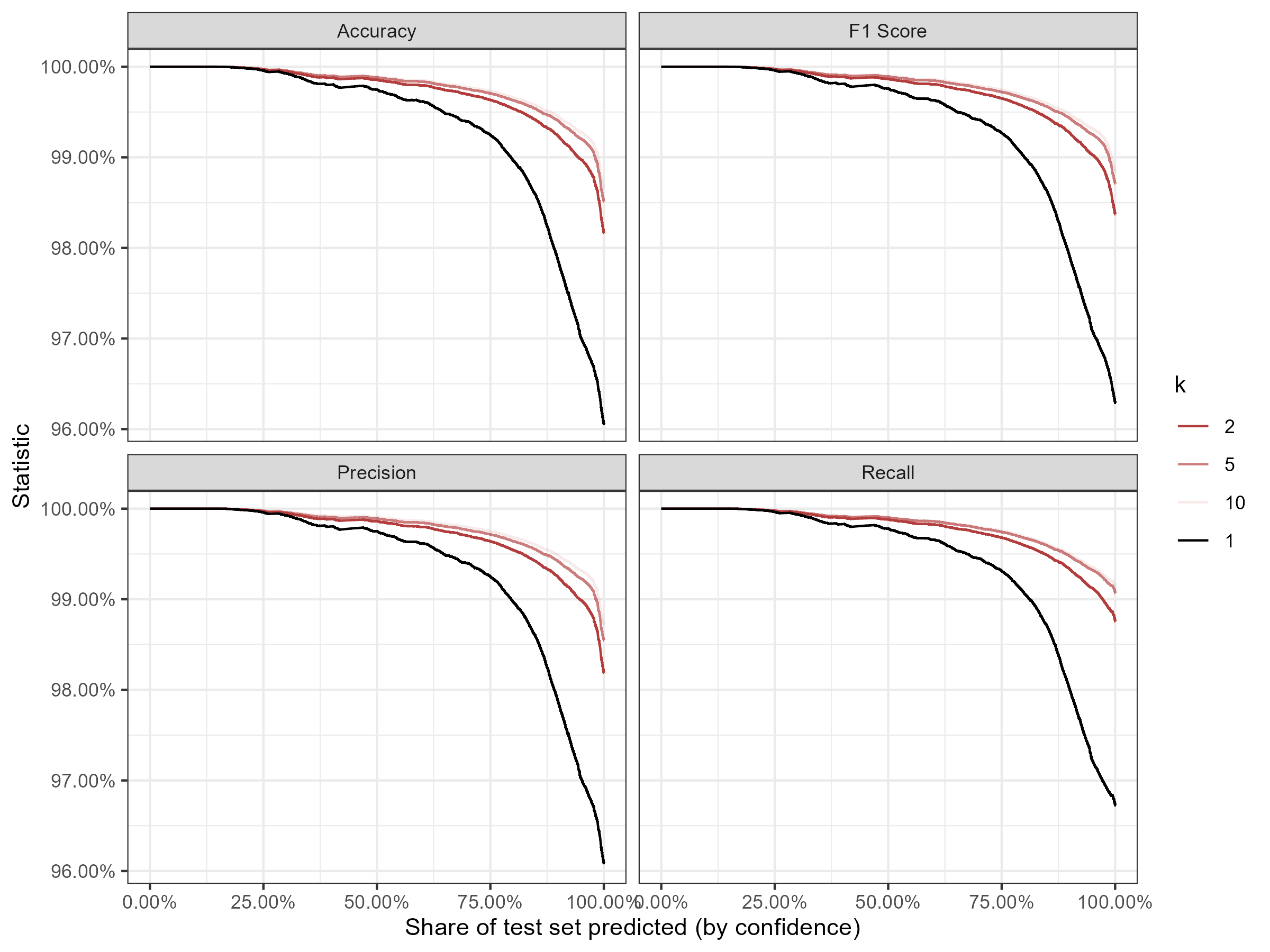}
    \parbox{0.9\textwidth}{
        \small \textit{Notes:} 
        The figure overlays production curves for greedy decoding (\(k=1\)) and top-\(k\) decoding. 
        Allowing a short list of candidate codes improves performance among predicted observations, especially in the low-confidence tail.\\
        \textit{Source}: Test data.
    }
    \label{fig:production_curve_topk}
\end{figure}

Top-\(k\) decoding modifies this picture in a simple way: it relaxes the requirement that the model must place the correct code at rank one. Figure~\ref{fig:production_curve_topk} overlays production curves that treat a prediction as correct if the true code is contained in the top-\(k\) candidates. The main takeaway is that the gain from top-\(k\) is not uniform: it is most valuable precisely where confidence is low and the top-ranked choice is fragile, but the correct code is still among the model's plausible alternatives.

\FloatBarrier

Table~\ref{tab:topk_avg_improvement} aggregates this idea across our main benchmarks. Figure~\ref{fig:topk_avg_improvement} in the main text provides a visual summary of the same average gains and diminishing returns as \(k\) increases. It reports average improvements from using a short shortlist rather than a single code, and it highlights the diminishing-returns pattern that motivates using small \(k\) in practice. In other words, the table is best read as guidance for choosing a default \(k\): it shows that a modest shortlist provides most of the benefit, while larger lists primarily matter when the application demands the last few percentage points of assisted-coding recall.

\begin{table}[h]
    \centering
    \caption{Average Performance Improvement Across \(k\)}
    \label{tab:topk_avg_improvement}
    \footnotesize
    \setlength{\tabcolsep}{6pt}
    \begin{tabular}[t]{lrrrr}
        \toprule
        Metric & Baseline ($k=1$) & $\Delta(k=2)$ & $\Delta(k=5)$ & $\Delta(k=10)$\\
        \midrule
        Accuracy & 0.961 & 0.021 & 0.025 & 0.026\\
        F1 score & 0.963 & 0.023 & 0.027 & 0.028\\
        Precision & 0.961 & 0.021 & 0.025 & 0.027\\
        Recall & 0.967 & 0.027 & 0.030 & 0.031\\
        \bottomrule
    \end{tabular}
    \parbox{0.85\textwidth}{
        \vspace{0.25cm}
        \footnotesize
        \textit{Notes}: Baseline is test-set performance (in percent) under greedy decoding (\(k=1\)). For \(k>1\), the decoder produces \(k\) candidate codes; \(\Delta(k)\) reports the increase in each metric relative to greedy decoding when evaluation uses the top-\(k\) candidate set.\\
        \textit{Source}: Test data.
    }
\end{table}

\FloatBarrier

The pooled averages conceal meaningful heterogeneity. Table~\ref{tab:topk_by_lang} reports the same comparison by language. The purpose is not to rank languages, but to document that the value of top-\(k\) depends on how close the evaluation data are to the model's training distribution and how standardized the occupational vocabulary is. In practical terms, the table motivates treating \(k\) as a tunable parameter rather than a fixed design choice: it can be chosen to match the application and the coder's time budget.

{\footnotesize
\begin{longtable}{l r l r r r r}
\caption{Top-\(k\) Performance Improvement by Language}
\label{tab:topk_by_lang}\\

\toprule
Language & Observations & Metric & Baseline & $\Delta(k=2)$ & $\Delta(k=5)$ & $\Delta(k=10)$\\
\midrule
\endfirsthead

\caption[]{Top-\(k\) Performance Improvement by Language (continued)}\\
\toprule
Language & Observations & Metric & Baseline & $\Delta(k=2)$ & $\Delta(k=5)$ & $\Delta(k=10)$\\
\midrule
\endhead

\midrule
\multicolumn{7}{r}{\footnotesize Continued on next page.}\\
\endfoot

\bottomrule
\multicolumn{7}{p{0.85\textwidth}}{\footnotesize \textit{Notes}: Baseline is test-set performance under greedy decoding (\(k=1\)). For \(k>1\), the decoder produces \(k\) candidate codes; \(\Delta(k)\) reports the increase in each metric relative to greedy decoding when evaluation uses the top-\(k\) candidate set.}\\
\multicolumn{7}{p{0.85\textwidth}}{\footnotesize \textit{Source}: Test data.}\\
\endlastfoot

\multirow{4}{*}{ca} & \multirow{4}{*}{330,290} & Accuracy & 0.998 & 0.001 & 0.002 & 0.002\\
& & F1 score & 0.998 & 0.002 & 0.002 & 0.002\\
& & Precision & 0.998 & 0.002 & 0.002 & 0.002\\
& & Recall & 0.998 & 0.002 & 0.002 & 0.002\\
\midrule
\multirow{4}{*}{da} & \multirow{4}{*}{2,745,570} & Accuracy & 0.983 & 0.007 & 0.008 & 0.010\\
& & F1 score & 0.986 & 0.009 & 0.010 & 0.012\\
& & Precision & 0.983 & 0.007 & 0.008 & 0.011\\
& & Recall & 0.991 & 0.014 & 0.014 & 0.015\\
\midrule
\multirow{4}{*}{en} & \multirow{4}{*}{3,787,830} & Accuracy & 0.944 & 0.036 & 0.042 & 0.043\\
& & F1 score & 0.944 & 0.036 & 0.042 & 0.043\\
& & Precision & 0.944 & 0.036 & 0.042 & 0.043\\
& & Recall & 0.944 & 0.036 & 0.042 & 0.043\\
\midrule
\multirow{4}{*}{es} & \multirow{4}{*}{4,270} & Accuracy & 0.966 & 0.032 & 0.032 & 0.032\\
& & F1 score & 0.966 & 0.032 & 0.032 & 0.032\\
& & Precision & 0.966 & 0.032 & 0.032 & 0.033\\
& & Recall & 0.966 & 0.032 & 0.032 & 0.032\\
\midrule
\multirow{4}{*}{fr} & \multirow{4}{*}{145,280} & Accuracy & 0.929 & 0.046 & 0.051 & 0.052\\
& & F1 score & 0.929 & 0.047 & 0.052 & 0.055\\
& & Precision & 0.929 & 0.047 & 0.055 & 0.062\\
& & Recall & 0.930 & 0.046 & 0.051 & 0.052\\
\midrule
\multirow{4}{*}{ge} & \multirow{4}{*}{67,200} & Accuracy & 0.900 & 0.058 & 0.063 & 0.064\\
& & F1 score & 0.901 & 0.060 & 0.066 & 0.068\\
& & Precision & 0.901 & 0.060 & 0.067 & 0.074\\
& & Recall & 0.904 & 0.061 & 0.066 & 0.067\\
\midrule
\multirow{4}{*}{gr} & \multirow{4}{*}{880} & Accuracy & 0.938 & 0.045 & 0.045 & 0.057\\
& & F1 score & 0.938 & 0.045 & 0.045 & 0.057\\
& & Precision & 0.938 & 0.045 & 0.045 & 0.057\\
& & Recall & 0.938 & 0.045 & 0.045 & 0.057\\
\midrule
\multirow{4}{*}{is} & \multirow{4}{*}{10,420} & Accuracy & 0.964 & 0.022 & 0.024 & 0.024\\
& & F1 score & 0.964 & 0.022 & 0.024 & 0.024\\
& & Precision & 0.964 & 0.022 & 0.024 & 0.024\\
& & Recall & 0.964 & 0.022 & 0.024 & 0.024\\
\midrule
\multirow{4}{*}{it} & \multirow{4}{*}{2,440} & Accuracy & 0.992 & 0.006 & 0.008 & 0.008\\
& & F1 score & 0.992 & 0.006 & 0.008 & 0.008\\
& & Precision & 0.992 & 0.006 & 0.008 & 0.008\\
& & Recall & 0.992 & 0.006 & 0.008 & 0.008\\
\midrule
\multirow{4}{*}{nl} & \multirow{4}{*}{595,360} & Accuracy & 0.980 & 0.009 & 0.012 & 0.012\\
& & F1 score & 0.980 & 0.010 & 0.012 & 0.013\\
& & Precision & 0.980 & 0.010 & 0.012 & 0.013\\
& & Recall & 0.981 & 0.010 & 0.012 & 0.013\\
\midrule
\multirow{4}{*}{no} & \multirow{4}{*}{81,880} & Accuracy & 0.987 & 0.008 & 0.010 & 0.010\\
& & F1 score & 0.987 & 0.008 & 0.010 & 0.010\\
& & Precision & 0.987 & 0.008 & 0.010 & 0.010\\
& & Recall & 0.987 & 0.008 & 0.010 & 0.010\\
\midrule
\multirow{4}{*}{pt} & \multirow{4}{*}{9,760} & Accuracy & 0.997 & 0.003 & 0.003 & 0.003\\
& & F1 score & 0.998 & 0.003 & 0.003 & 0.003\\
& & Precision & 0.997 & 0.003 & 0.003 & 0.003\\
& & Recall & 0.998 & 0.003 & 0.003 & 0.003\\
\midrule
\multirow{4}{*}{se} & \multirow{4}{*}{1,118,770} & Accuracy & 0.941 & 0.015 & 0.020 & 0.022\\
& & F1 score & 0.952 & 0.026 & 0.030 & 0.032\\
& & Precision & 0.941 & 0.016 & 0.020 & 0.023\\
& & Recall & 0.973 & 0.047 & 0.050 & 0.050\\
\midrule
\multirow{4}{*}{unk} & \multirow{4}{*}{414,790} & Accuracy & 0.980 & 0.017 & 0.017 & 0.017\\
& & F1 score & 0.980 & 0.017 & 0.017 & 0.017\\
& & Precision & 0.980 & 0.017 & 0.017 & 0.017\\
& & Recall & 0.980 & 0.017 & 0.017 & 0.017\\

\end{longtable}
}

\FloatBarrier

Finally, we evaluate top-\(k\) decoding on out-of-distribution datasets. The purpose of Table~\ref{tab:topk_ood} is to illustrate a common deployment scenario. Performance might be poor. But a short shortlist often contains the correct code and therefore meaningfully reduces human labeling time. 

\begin{table}[h]
    \centering
    \caption{Top-\(k\) Performance Improvement for Out-of-distribution Datasets}
    \label{tab:topk_ood}
    \footnotesize
    \begin{tabular}[t]{l r l r r r r}
        \toprule
        Dataset & Observations & Metric & Baseline & $\Delta(k=2)$ & $\Delta(k=5)$ & $\Delta(k=10)$\\
        \midrule
        \multicolumn{7}{l}{\textit{Panel A: Raw data}}\\
        \midrule
        \multirow{4}{*}{Swedish Strikes} & \multirow{4}{*}{1,430} & Accuracy & 0.794 & 0.064 & 0.071 & 0.075\\
        & & F1 score & 0.795 & 0.065 & 0.072 & 0.077\\
        & & Precision & 0.794 & 0.064 & 0.071 & 0.075\\
        & & Recall & 0.797 & 0.067 & 0.073 & 0.080\\
        \midrule
        \multirow{4}{*}{Dutch Wealth Tax} & \multirow{4}{*}{200} & Accuracy & 0.520 & 0.365 & 0.380 & 0.390\\
        & & F1 score & 0.520 & 0.365 & 0.380 & 0.390\\
        & & Precision & 0.520 & 0.365 & 0.380 & 0.390\\
        & & Recall & 0.520 & 0.365 & 0.380 & 0.390\\
        \midrule
        \multirow{4}{*}{German Denazification Survey} & \multirow{4}{*}{800} & Accuracy & 0.329 & 0.066 & 0.118 & 0.158\\
        & & F1 score & 0.342 & 0.077 & 0.129 & 0.166\\
        & & Precision & 0.329 & 0.066 & 0.118 & 0.158\\
        & & Recall & 0.368 & 0.101 & 0.151 & 0.182\\
        \midrule
        \multirow{4}{*}{Danish West Indies} & \multirow{4}{*}{166,563} & Accuracy & 0.291 & 0.065 & 0.411 & 0.431\\
        & & F1 score & 0.291 & 0.065 & 0.411 & 0.431\\
        & & Precision & 0.291 & 0.065 & 0.411 & 0.431\\
        & & Recall & 0.291 & 0.065 & 0.411 & 0.431\\
        \midrule
        \multirow{4}{*}{UK Bankruptcies} & \multirow{4}{*}{581,912} & Accuracy & 0.277 & 0.075 & 0.137 & 0.209\\
        & & F1 score & 0.277 & 0.075 & 0.137 & 0.209\\
        & & Precision & 0.277 & 0.075 & 0.137 & 0.209\\
        & & Recall & 0.277 & 0.075 & 0.137 & 0.209\\
        \midrule
        \multicolumn{7}{l}{\textit{Panel B: Removed HISCO code '-1': No occupation}}\\
        \midrule
        \multirow{4}{*}{Danish West Indies} & \multirow{4}{*}{101,619} & Accuracy & 0.465 & 0.073 & 0.128 & 0.128\\
        & & F1 score & 0.465 & 0.073 & 0.128 & 0.128\\
        & & Precision & 0.465 & 0.073 & 0.128 & 0.128\\
        & & Recall & 0.465 & 0.073 & 0.128 & 0.128\\
        \midrule
        \multirow{4}{*}{UK Bankruptcies} & \multirow{4}{*}{326,400} & Accuracy & 0.440 & 0.118 & 0.125 & 0.125\\
        & & F1 score & 0.440 & 0.118 & 0.125 & 0.125\\
        & & Precision & 0.440 & 0.118 & 0.125 & 0.125\\
        & & Recall & 0.440 & 0.118 & 0.125 & 0.125\\
        \bottomrule
    \end{tabular}
    \parbox{0.85\textwidth}{
        \vspace{0.25cm}
        \footnotesize
        \textit{Notes}: 
        Baseline shows \(k=1\) performance; \(\Delta(k)\) shows improvement over baseline.
        Baseline is performance under greedy decoding (\(k=1\)). For \(k>1\), the decoder produces \(k\) candidate codes; \(\Delta(k)\) reports the increase in each metric relative to greedy decoding when evaluation uses the top-\(k\) candidate set. Panel B removes HISCO code `-1' (\textit{no occupation}) before evaluation.\\
        \textit{Source}: Out-of-distribution evaluation sets; dataset descriptions and citations are reported in Table~\ref{tab:ood-testing-automatic}.
    }
\end{table}

\begin{figure}
    \centering
    \caption{Out-of-domain Top-\(k\) Performance on Danish West Indies Data}
    \vspace{0.5cm}
    \includegraphics[width=\linewidth]{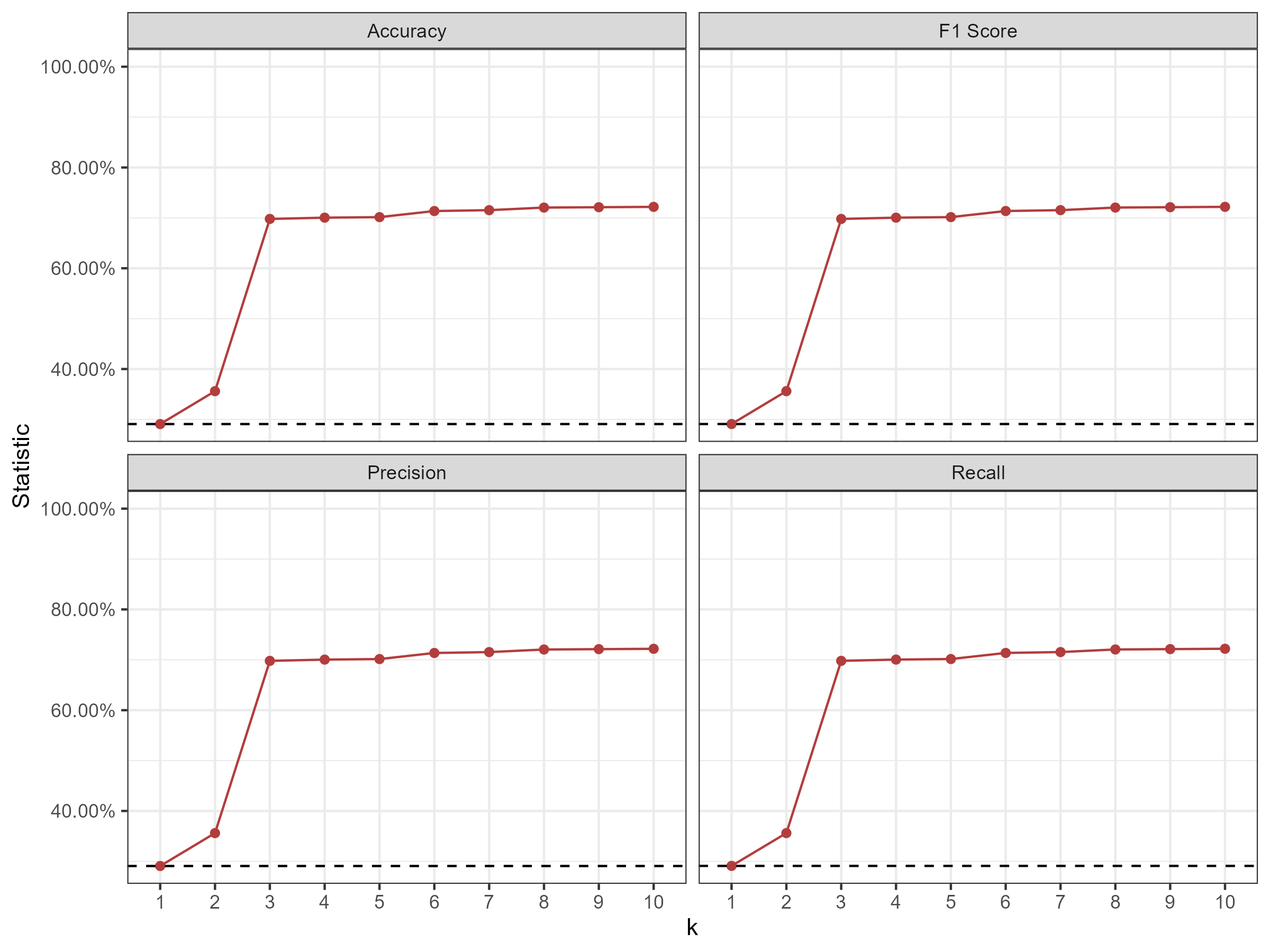}
    \parbox{0.9\textwidth}{
        \small \textit{Notes:} 
        The figure shows out-of-domain model performance on the Danish West Indies out-of-domain data.
        Out-of-domain performance can be low under greedy decoding, but a short candidate list can recover many ``near misses'' in assisted-coding settings. \\
        \textit{Source}: Out-of-domain evaluation set.
    }
    \label{fig:topk_ood_example}
\end{figure}

Figure~\ref{fig:topk_ood_example} provides a visual illustration of the same mechanism on one dataset. top-\(k\) decoding can substantially improve performance. 

\FloatBarrier
\section{How OccCANINE understands occupations} \label{sec:embeddings}
\begin{figure}[ht]
    \centering
    \caption{t-SNE Visualization of Occupational Embeddings}
    \includegraphics[width=\textwidth]{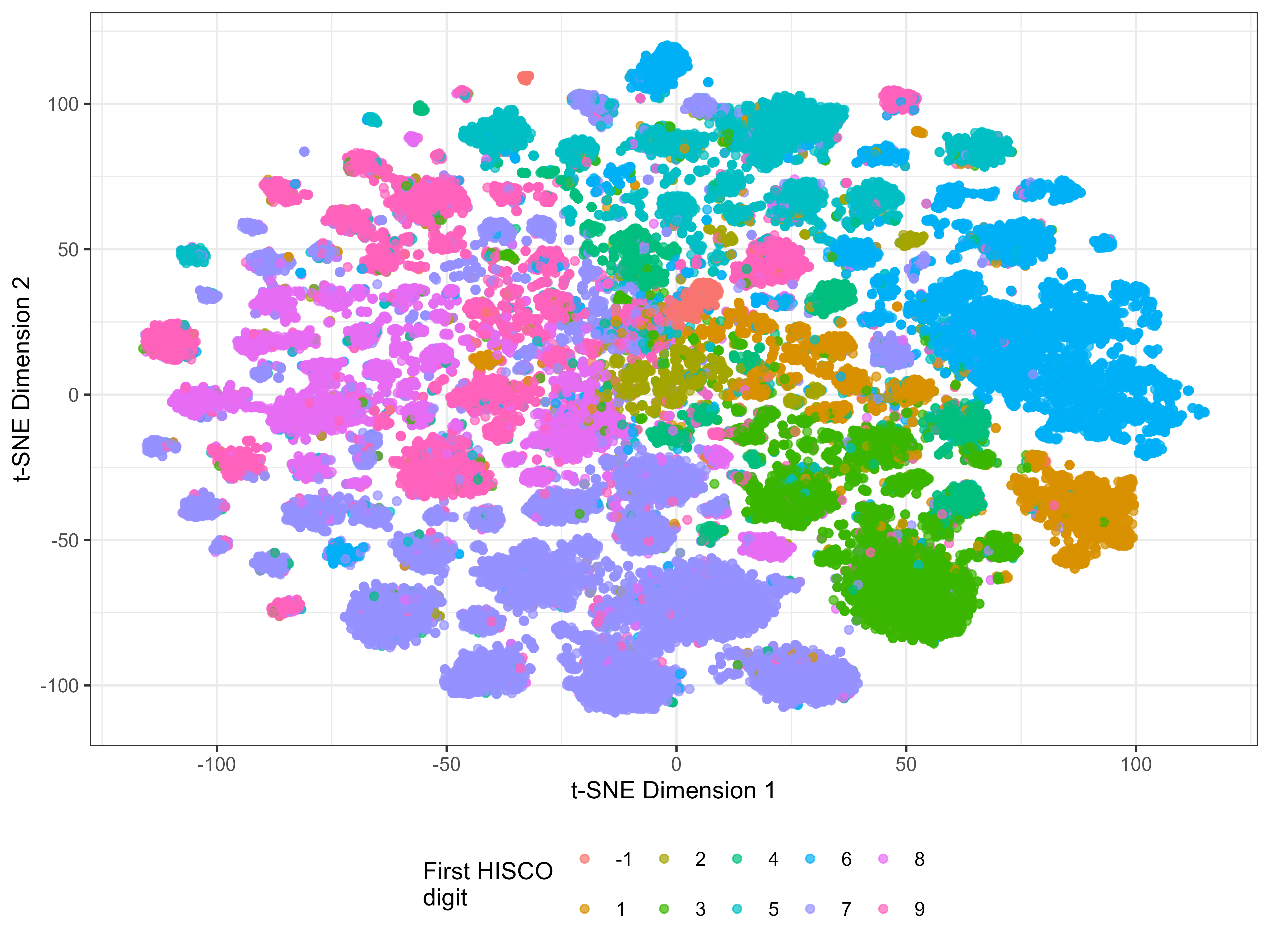}
    \parbox{0.9\textwidth}{
        \vspace{0.25cm}
        \footnotesize \textit{Notes}: The figure shows embeddings from  \textit{OccCANINE}. The colors correspond to the first digit of the HISCO code, roughly indicative of economic sectors. The data depicted was not seen during model training. \\
        \textit{Source}: 100,000 test observations (downsampled to avoid clutter).
        }
    \label{fig:tsne_visualization}
\end{figure}

To investigate the underlying semantic knowledge that the model has obtained, we randomly sample embeddings for 100,000 validation observations.\footnote{The output from the final layer of the encoder.} This 768 dimensional output represents the meaning of each occupational description in a continuous space. If similar descriptions cluster closely, this suggests that the model has acquired a structural comprehension of occupations. Figure~\ref{fig:tsne_visualization} shows a low-dimensional representation of these embeddings using t-SNE \citep{maaten2008visualizing}. The colors represent the first digit of the HISCO code. This roughly represents different sectors of the economy.  Note how occupations which are closer together tend to have the same color. This suggests that \textit{OccCANINE} picks up similar occupations as being semantically similar. This result shows the potential for generalized high performance across different domains, and in turn, it suggests that \textit{OccCANINE} is a valuable starting point for other applications related to occupational descriptions in historical settings.

\FloatBarrier
\section{Evaluation Metrics}
\label{sec:performance_measures}

Evaluating occupational coding models requires metrics that handle several complications: (i) both ground truth and predictions can contain multiple valid codes per observation; (ii) codes may vary in length, and partially correct predictions (matching leading digits) still carry information; and (iii) duplicates, missing values, and empty strings should not influence the comparison. To address this, we adapt standard classification measures to our setting.  

We first define each metric at the level of an individual observation, comparing its set of true codes $Y_i$ with the predicted codes $\hat{Y}_i$. Once these per-observation scores are obtained, dataset-level metrics are simply the average across all observations, giving each case equal weight regardless of the number of associated codes. If truncation to $d$ digits is specified, codes are truncated before comparison.

Accuracy is defined as
\begin{equation}
\text{Acc}(Y_i, \hat{Y}_i) = 
\frac{1}{2 \cdot \max(|Y_i|, |\hat{Y}_i|)}
\left(
\sum_{y \in \hat{Y}_i} \mathbf{1}\{y \in Y_i\} +
\sum_{y \in Y_i} \mathbf{1}\{y \in \hat{Y}_i\}
\right).
\end{equation}
This measure averages two directions of overlap: predictions found among the true codes, and true codes found among the predictions. Dividing by the maximum set size ensures balance when the sets differ in size. Unlike the classical definition of accuracy, $(TP+TN)/N$, this formulation does not rely on true negatives, which are not meaningful in a setting with multiple possible occupational codes.

Precision is defined as
\begin{equation}
\text{Prec}(Y_i, \hat{Y}_i) =
\begin{cases}
0 & \text{if } |\hat{Y}_i| = 0, \\[6pt]
\dfrac{\sum_{y \in \hat{Y}_i} \mathbf{1}\{y \in Y_i\}}{|\hat{Y}_i|} & \text{otherwise.}
\end{cases}
\end{equation}
It measures the fraction of predicted codes that are correct. This corresponds directly to the classical definition, $\tfrac{TP}{TP+FP}$, with “true positives” and “false positives” understood as set membership of occupational codes.

Recall is defined as
\begin{equation}
\text{Rec}(Y_i, \hat{Y}_i) =
\begin{cases}
0 & \text{if } |Y_i| = 0, \\[6pt]
\dfrac{\sum_{y \in Y_i} \mathbf{1}\{y \in \hat{Y}_i\}}{|Y_i|} & \text{otherwise.}
\end{cases}
\end{equation}
It measures the fraction of true codes recovered by the model. This matches the classical definition, $\tfrac{TP}{TP+FN}$, applied to sets of occupational codes rather than binary outcomes.

Finally, the F1 score is defined as
\begin{equation}
\text{F1}(Y_i, \hat{Y}_i) =
\begin{cases}
0 & \text{if } \text{Prec}(Y_i, \hat{Y}_i) + \text{Rec}(Y_i, \hat{Y}_i) = 0, \\[6pt]
\dfrac{2 \cdot \text{Prec}(Y_i,\hat{Y}_i) \cdot \text{Rec}(Y_i,\hat{Y}_i)}
{\text{Prec}(Y_i,\hat{Y}_i) + \text{Rec}(Y_i,\hat{Y}_i)} & \text{otherwise.}
\end{cases}
\end{equation}
This balances precision and recall by taking their harmonic mean. It is identical to the textbook definition, but here the underlying precision and recall are adapted to the multi-label, variable-length occupational coding problem. That is, it measures the per-example set metrics. 

\section{Data processing pipeline}\label{app:data_processing}

Because our training data come from heterogeneous projects and archival contexts, we apply a semi-standardized processing pipeline across all sources. The aim is to remove formatting differences while preserving meaningful variation in how occupations are recorded. In practice, this pipeline has four components.

First, we perform \textbf{character normalization}, replacing non-English characters with standard English equivalents (e.g., ``æ'' becomes ``ae'', ``ø'' becomes ``oe'', etc.). This step mainly prevents the same occupational string from appearing in multiple encodings and helps the model treat superficial orthographic variation consistently. Second, we conduct \textbf{source-specific checks} via manual inspection to identify idiosyncrasies in each dataset. For example, some sources provide both a ``raw'' and a ``clean'' occupational description; in such cases, we retain both, since each is a plausible input format and including both improves coverage.

Third, we apply \textbf{HISCO validation} and retain only observations with standard HISCO codes, dropping cases with non-standard or modified codes. This matters because even small deviations in coding practice can make labels incomparable across sources. As a concrete example, IPUMS (\citealp{IPUMS}) provides its own adaptation of HISCO; we therefore restrict attention to cases where codes were unaltered from the original standard, based on the crosswalk provided by \cite{ipums_cross_walk2017}. Finally, we generate \textbf{multi-occupation phrasing} by combining two single-occupation strings with an appropriate conjunction in the same language. This is intended to teach the model to report \emph{all} relevant HISCO codes when an occupational description refers to multiple occupations rather than implicitly selecting one; see Section~\ref{app:conjunctions} for details.

\FloatBarrier
\section{Conjunction augmentation}\label{app:conjunctions}

We augment all data \emph{within each data source} by creating two-occupation strings in the same language, combining a base description with another from that source using the appropriate conjunction (e.g., \textit{og/and/und/en/och}). Each base record is paired with up to $10$ partners to expand coverage of realistic multi-occupation phrases without overwhelming the original distribution. 

For very large sources, we first draw a simple random sample of at most $10{,}000$ single-occupation records before forming pairs to keep data growth in check. The procedure is applied consistently across sources and therefore carries through to all derived splits, with a fixed random seed ensuring reproducibility.

\section{Training details}\label{app:training_details}

Training ran for 1,605,000 steps with a batch size of 512. In the training procedure, the language is provided first, followed by a separator and then the occupational description. \textit{CANINE} applies 10 percent dropout between all layers by default, which we keep unchanged.

We regularize training with robustness augmentation in two stages. First, \emph{before} training, we expand the training set with a fixed pool of hard examples produced using an earlier version of the model as a ``probe.'' Starting from the original data, we generate adversarial variants of occupational strings (including character- and word-level edits and back-and-forth machine translation) and retain variants that alter the earlier model's prediction. In addition, we include machine-translated versions of occupational descriptions across the training languages and randomly generated letter/space strings labelled as ``no occupation'' (-1).\footnote{Translations performed using \cite{fan2020translation}} The translated strings broaden linguistic coverage, while the random strings discourage the model from producing plausible-looking codes when the input is nonsensical.

Second, \emph{during} training, we apply on-the-fly perturbations in the data loader so that the model continually sees slightly different surface forms across steps. These transformations are inspired by TextAttack \citep{morris2020textattack}: each input has a 10 percent chance of random word insertion and a separate 10 percent chance of random character alteration, with each character then having a 10 percent chance of being replaced by a random character. To improve robustness when language metadata are missing, we also randomly set the language for an observation to ``unk'' (for \textit{unknown}) with a 25 percent probability.

\FloatBarrier
\section{Order-invariant loss function}
\label{app:order-invariant-loss}

Generally speaking, the occupational codes in our labels are not consistently ordered across training samples. This is not an inherent issue, as our objective is to detect all occupations mentioned within a text, regardless of their order. However, a challenge of using a sequence-to-sequence model is how to best represent the labels in such a case. The classical cross entropy loss as used when training, e.g., other transformer models will not be optimal, as it will punish the model when outputting the occupational codes in the ``wrong'' order -- despite each permutation being equally valid.

To see why, consider the following example. A description such as ``he fishes and farms'' may correspond to the codes \texttt{61110 General Farmer} and \texttt{64100 Fisherman} in any order; that is, in our training data, this may be represented as either of
\begin{align}
    \texttt{BOS, } \underbrace{\texttt{6, 1, 1, 1, 0}}_{\text{Farmer}} \texttt{, } \underbrace{\texttt{6, 4, 1, 0, 0}}_{\text{Fisher}} \texttt{, } \underbrace{\dots}_{\text{Other codes}} \texttt{, EOS} &,
    \\
    \texttt{BOS, } \underbrace{\texttt{6, 4, 1, 0, 0}}_{\text{Fisher}} \texttt{, } \underbrace{\texttt{6, 1, 1, 1, 0}}_{\text{Farmer}} \texttt{, } \underbrace{\dots}_{\text{Other codes}} \texttt{, EOS} &,
\end{align}
where \texttt{BOS} denotes a special ``beginning of sequence'' and \texttt{EOS} a special ``end of sequence'' token.

Suppose now that our model, based on the input string ``he fishes and farms'', predicts the sequence as \texttt{BOS, 6, 1, 1, 0, 6, 4, 1, 0, 0, \dots, EOS}.\footnote{More precisely, the model will predict a probability distribution over the tokens for each element in the sequence, which in the extreme case may put mass only at exactly these tokens.} In case (5) above, where this exactly matches the label, this leads to a perfect prediction with loss 0. In case (6), however, this leads to a score no higher than gibberish, when in fact we view it as equally valid.

To address this, we propose a loss function that is invariant to the order of the predicted codes -- which we term the ``blocks'' of the prediction. The idea of our ``block order-invariant classification loss'' is to select, for each non-empty target block, the best matching candidate block without imposing any (arbitrary) order.

\subsection{Block order-invariant classification loss}

Let the network output be represented as a tensor of logits $\hat{\boldsymbol{Y}} \in \mathbb{R}^{(N \cdot b) \times V}$, where $N$ is the number of blocks, $b$ is the size of each block, and $V$ is the vocabulary size.
Similarly, let the target sequence be given by $\boldsymbol{Y} \in \mathbb{N}_0^{(N \cdot b)}$, where a special padding token $p$ is used to denote empty positions. 

We now partition both the target and candidate output into $N$ blocks of length $b$.
Denote these by
\begin{align*}
    \hat{\boldsymbol{Y}}_i &= \hat{\boldsymbol{Y}}[ib:(i + 1)b] \in \mathbb{R}^{b \times V}, \quad \text{for } i=0, \dots, N-1,
    \\
    \boldsymbol{Y}_j &= \boldsymbol{Y}[jb:(j + 1)b] \in \mathbb{N}_0^{b}, \quad \text{for } j=0, \dots, N-1,
\end{align*}
where the notation $\hat{\boldsymbol{Y}}[i_1:i_2]$ means the subsequence from position $i_1$ up to, but excluding, position $i_2$ (i.e., of length $i_2 - i_1$).
Note that we index from 0.

For each pair $ij$ of candidate ($i$) and target ($j$) blocks, we compute the cross entropy (CE) loss in the ordinary element-wise fashion
\begin{align*}
    L_{ij} = \frac{1}{b} \sum_{\ell = 0}^{b - 1} \text{CE}\left( \hat{\boldsymbol{Y}}_i[\ell], \boldsymbol{Y}_j[\ell] \right),
\end{align*}
where $\hat{\boldsymbol{Y}}_i[\ell]$ denotes the $\ell$th element of the $i$th block.

Let $k \leq N$ denote the number of valid (i.e., non-padded) target blocks.
To achieve order-invariance \textit{across} blocks, we aggregate these pair-wise losses as\footnote{When applied on a batch, rather than a single observation, we divide by the smallest value $k^*$ such that any $k$ is no greater than $k^*$.}
\begin{align*}
    L_{\text{inv}} = \frac{1}{k} \sum_{j = 0}^{k - 1} \, \min_{0 \leq i < k} \, L_{ij}
\end{align*}

\paragraph{Inference-time differences}

Since we train our model with teacher-forcing, our model can partially ``cheat'' on sequences with $K > 1$.
Specifically, it can, in principle, look at the correct values of the first code when predicting its second block, and due to order invariance, these two can then be matched against each other during loss calculation.

During inference, where we use a greedy, autoregressive decoding scheme, this cannot happen, and our experiments show that the above does not lead to any undesirable behavior.
For example, our accuracy is more or less similar regardless of whether decoding autoregressively or using teacher forcing.


\subsection{Promoting sparsity}

Relying only on $L_{\text{inv}}$ has the limitation of not encouraging ``sparsity'' in predictions, i.e., it is necessary to discourage the network from producing extra, spurious predictions.\footnote{This happens since no penalty is applied to producing more predictions than there are valid target blocks.}
For this purpose, we apply a sparsity-promoting loss to candidate blocks that exceed the number of valid target blocks.
Define the binary mask $\mathcal{M} \in \{0, 1\}^N$ as
\begin{align*}
    \mathcal{M}_i = \begin{cases}
        1, & \text{if block } i \text{ is not associated with any target (i.e., } i \geq k\text{)},\\[1mm]
        0, & \text{otherwise.}
    \end{cases}
\end{align*}

We next compute, for each candidate block $\hat{\boldsymbol{Y}}_j$, a padding loss (which is simply cross entropy letting all targets take the padding value $p$)
\begin{align*}
    L_{\text{pad}}^j = \frac{1}{b} \sum_{\ell = 0}^{b - 1} \text{CE}\left(\hat{\boldsymbol{Y}}_{j_{\ell}}, p\right)
\end{align*}

Finally, we compute our sparsity promoting loss as
\begin{align*}
    L_{\text{pad}} = \frac{1}{N} \sum_{i = 0}^{N - 1} \mathcal{M}_i L_{\text{pad}}^i
\end{align*}

\subsection{Computational implementation}

To efficiently compute our loss across a batch of observations, we make use of 0- and $\infty$-masking to allow us to iterate over the same number of blocks for all observations within a batch.
We use $0$-masking to handle terms in our summation beyond our desired target and $\infty$-masking to handle appropriately selecting the minima when computing $L_{\text{inv}}$.
Mathematically, this is equivalent, but is much faster due to added parallelization.


\subsection{Total loss}

The complete loss function combines the order-invariant block loss and the sparsity loss with an optional scaling factor $\lambda$
\begin{equation}
    L = L_{\mathrm{inv}} + \lambda \, L_{\mathrm{pad}},
\end{equation}
where $\lambda$ is a hyperparameter, balancing the contribution of the sparsity-enforcing term relative to the block classification loss.
We let $\lambda = 1$ in all our experiments.


\section{Loss mixer}

To train models to simultaneously produce predictions based both on a transformer decoder and a linear classifier, we use a loss mixer.
Section~\ref{app:order-invariant-loss} covers the case for the transformer decoder.
For our linear decoder, we use binary cross entropy (BCE).

Let $M$ denote the number of codes, and let the network's linear output be represented as a tensor of logits $\hat{\boldsymbol{Z}} \in \mathbb{R}^M$ and the linear target be given by $\boldsymbol{Z} \in \{0, 1\}^M$.
The loss for the linear output is then
\begin{align*}
    L_{\text{Linear}} = \frac{1}{M} \sum_{i = 0}^{M - 1}  \text{BCE}(\hat{\boldsymbol{Z}}_i, \boldsymbol{Z}_i),
\end{align*}
where $\hat{\boldsymbol{Z}}_i$ denotes the $i$th element of the linear output.

Combining our order invariant loss with this linear loss, our mixer loss function is given by
\begin{align*}
    L_{\text{Mixer}} = \gamma L + (1 - \gamma) L_{\text{Linear}},
\end{align*}
where $\gamma \in (0, 1)$ is a hyperparameter that controls the relative strength of our order invariant loss to our linear loss.
Since the absolute magnitude of $L$ is substantially larger than $L_{\text{Linear}}$, we found it beneficial to use low values of $\gamma$, to ensure our linear decoder also obtains satisfactory performance.
In all reported results, we use $\gamma = 0.1$.

\end{appendices}